\documentclass[review]{elsarticle}

\usepackage{lineno,hyperref}
\modulolinenumbers[5]

\usepackage[nolist]{acronym}
\begin{acronym}
\acro{fnn}[FNN]{Feed-Forward Neural Network}
\acro{nsgaiii}[NSGA-III]{Non-dominated Sorting Genetic Algorithm III}
\end{acronym}    
\usepackage{amsmath,amsthm,amssymb,epsfig,bm}
\usepackage{array}
\usepackage{float}

\usepackage{framed}
\usepackage{graphicx}
\usepackage{cleveref}
\usepackage{float}
\usepackage{footnote}
\usepackage{comment}
\usepackage{subcaption}
\usepackage[yyyymmdd]{datetime}
\usepackage{booktabs,caption}
\usepackage[]{hyperref}
\usepackage{orcidlink}
\hypersetup{
 colorlinks  = true, 
 urlcolor   = blue, 
 linkcolor  = black, 
 citecolor  = blue 
}
\usepackage[utf8]{inputenc}
 
\usepackage{listings}
\usepackage{color}
\definecolor{codegreen}{rgb}{0,0.6,0}
\definecolor{codegray}{rgb}{0.5,0.5,0.5}
\definecolor{codepurple}{rgb}{0.58,0,0.82}
\definecolor{backcolour}{rgb}{0.95,0.95,0.92}
\lstdefinestyle{mystyle}{
  backgroundcolor=\color{backcolour},  
  commentstyle=\color{codegreen},
  keywordstyle=\color{magenta},
  numberstyle=\tiny\color{codegray},
  stringstyle=\color{codepurple},
  basicstyle=\footnotesize,
  breakatwhitespace=false,     
  breaklines=true,         
  captionpos=b,          
  keepspaces=true,         
  numbers=left,          
  numbersep=5pt,         
  showspaces=false,        
  showstringspaces=false,
  showtabs=false,         
  tabsize=2,
  escapeinside={<@}{@>},
}
 
\usepackage{booktabs} 

\usepackage{soul} 
\soulregister\cite7
\soulregister\ref7
\soulregister\pageref7
\usepackage{bm}

\usepackage{nomencl} 
\usepackage{tikz}
\makenomenclature

\usepackage{multicol}

\journal{Swarm and Evolutionary Computation}

\begin{document}

\begin{frontmatter}


\title{\large MAEO: Multiobjective Animorphic Ensemble Optimization for Scalable Large-scale Engineering Applications}



\author{Omer F. Erdem$^{a*}$ \orcidlink{0000-0002-2931-3591}, Dean Price$^{c}$ \orcidlink{0000-0003-0999-0111}, Paul Seurin$^{d}$ \orcidlink{0000-0002-5940-7695}, Majdi I. Radaideh$^{a,b*}$ \orcidlink{0000-0002-2743-0567}}

\cortext[mycorrespondingauthor]{Corresponding Author: M. I. Radaideh (radaideh@umich.edu), Omer F. Erdem (oferdem@umich.edu)}

\address{$^{a}$Department of Nuclear Engineering and Radiological Sciences, University of Michigan, Ann Arbor, MI 48109, United States}

\address{$^{b}$Department of Computer Science and Engineering, University of Michigan, Ann Arbor, MI 48109, United States}

\address{$^{c}$Department of Nuclear Science and Engineering, Massachusetts Institute of Technology, 60 Vassar St., Cambridge, MA 02139}

\address{$^{d}$Idaho National Laboratory, Idaho Falls, ID 83415, United States}

\small

\begin{abstract}

Multiobjective optimization remains challenging for many scientific and engineering problems due to the need to balance convergence, diversity, and computational efficiency across high-dimensional objective landscapes. This work presents the \textbf{Multiobjective Animorphic Ensemble Optimization (MAEO)} framework, a parallelizable ensemble strategy that unifies state-of-the-art evolutionary algorithms within an island-based architecture, overcoming the limitations of relying on a single optimizer, as implied by the No Free Lunch theorem. MAEO uses a parameter-free hypervolume indicator for island performance assessment and a strict Pareto-rank–based individual scoring formulation that incorporates crowding distance and nadir-point proximity to ensure consistent selection pressure within each front. The framework is initiated using four algorithms (NSGA-III, CTAEA, AGEMOEA2, SPEA2) and evaluated through extensive benchmarking on 12 DTLZ/ZDT functions under 36 dimensionality settings using Wilcoxon signed-rank tests with both hypervolume and inverse generational distance metrics. Results show that MAEO achieves balanced convergence–diversity performance, outperforming or matching some of the leading multiobjective optimization algorithms across different benchmark problems. To demonstrate practical applicability, MAEO is applied to the equilibrium-cycle optimization of a small modular nuclear reactor. Eight discrete design variables (seven assembly enrichments and burnable absorber configuration) and three objectives (levelized cost of electricity, peak soluble boron concentration, fuel cycle length) are optimized under two safety constraints. The algorithm carried out roughly 40,000 evaluations using computer simulations. MAEO identifies core designs that lower both the levelized cost of electricity and the peak boron concentration, while preserving fuel cycle length and meeting all safety constraints. These findings highlight MAEO’s effectiveness for high-dimensional and large-scale engineering optimization.

\end{abstract}


\end{frontmatter}

\section{Introduction}
\label{sec:intro}

Optimization problems are common across all engineering disciplines \cite{hoghoj2023simultaneous}. They arise in the design of aircraft wings \cite{conlanSmith2020aerodynamic}, power systems \cite{jeong2021maximization, wilding2020use}, fuel management \cite{dechaine1996fuel}, bridge loading \cite{zaheer2022literature}, and many other applications. Unlike purely mathematical optimization problems, engineering optimization problems often involve high-cost fitness functions that are computationally expensive to evaluate. Moreover, they are typically treated as black-box problems, where the optimizer has no access to derivative information. As a result, solving such problems may require extensive computational time, often necessitating the use of high-performance computing clusters.

A wide range of modern, gradient-free methods have been developed for black-box optimization. These include direct search methods, population-based (evolutionary and swarm-based) methods, surrogate-assisted methods and hybrid approaches \cite{conn2009introduction}. Many of the population-based methods are nature-inspired, such as Genetic Algorithm (GA) \cite{holland1975adaptation}, Evolution Strategies (ES) \cite{back1996evolutionary}, Particle Swarm Optimization (PSO) \cite{kennedy1995particle}, Grey Wolf Optimization (GWO) \cite{mirjalili2014grey}, Moth-Flame Optimization (MFO) \cite{mirjalili2015moth}, Whale Optimization (WOA) \cite{mirjalili2016whale}, and JAYA \cite{ray2016jaya}. These methods have been extensively tested on a variety of benchmark problems, and their effectiveness has been demonstrated in numerous engineering studies \cite{jeong2021maximization, radaideh2022model, zhang2015multiobjective, seurin2024multi, erdem2025multi}. A wide range of Python and cross-language libraries has been developed to make evolutionary and swarm algorithms easily accessible, including DEAP (Distributed Evolutionary Algorithms in Python) \cite{fortin2012deap}, pyMOO (Multi-objective Optimization in Python) \cite{blank2020pymoo}, and NEORL (NeuroEvolution Optimisation with Reinforcement Learning) \cite{radaideh2023neorl,radaideh2021neorl}.

Reinforcement learning (RL) is widely used for optimization by framing design or control tasks as sequential decision-making problems, in which an agent interacts with an environment to learn policies that optimize an objective by maximizing long-term reward subject to constraints \cite{SeurinShirvan2024FuelOptimization,radaideh2025multistep}. Radaideh and Shirvan \cite{radaideh2021rule} introduced a rule-based reinforcement learning framework to guide evolutionary algorithms in handling complex constraints in engineering optimization problems. Physics-informed RL was introduced as an optimization paradigm in which physical laws are explicitly embedded into the learning process, ensuring physically consistent and data-efficient optimization for large-scale engineering systems, and was demonstrated for nuclear power applications at both small \cite{radaideh2021physics} and large scales \cite{seurin2024physics,radaideh2021large}.

However, there is no universal consensus on which method performs best for which type of problem, i.e., No Free Lunch theorem (NFL) \cite{wolpert1997no}. Consequently, the common practice in the engineering optimization literature is to apply multiple methods to the same problem and compare their outcomes to identify the most effective approach \cite{seurin2025surpassing, zhang2024ensemble, ye2024ensemble, rezk2024metaheuristic}. According to the NFL theorem, there is no single optimization algorithm that performs best across all problems \cite{wolpert1997no}. As a result, mathematicians and engineers continue to develop new algorithms that approach benchmark problems from different perspectives. To improve reliability and robustness, these algorithms are often used in parallel or combined within hybrid frameworks. In hybrid approaches, different search algorithms—such as Follow the Leader (FTL) \cite{yang2007ftl}, Multi-Verse Optimizer (MVO) \cite{mirjalili2016mvo}, and Salp Swarm Algorithm (SSA) \cite{mirjalili2017ssa}, or Prioritized Experience Replay for Parallel Hybrid Evolutionary and Swarm Algorithms (PESA) \cite{radaideh2022pesa}—are combined to leverage their complementary strengths. Adaptive strategies introduce mechanisms to evaluate sub-swarm performance dynamically and redistribute resources or re-divide populations according to the most suitable algorithms \cite{singh2021ensemble}. 

In multi-swarm methods, a population is divided into sub-swarms, with each sub-swarm using a different algorithm or strategy, as implemented in Multi-swarm particle swarm optimization (MPSO) \cite{jie2010multi} and Animorphic Ensemble Optimization (AEO) \cite{price2023animorphic}. These ensemble-based optimization algorithms combine multiple search processes to exploit complementary strengths, thereby improving robustness, convergence, and solution quality beyond what a single optimizer can achieve. These methods typically maintain multiple candidate populations or solvers in parallel, with interactions such as migration, information sharing, or adaptive weighting. Ensemble approaches are generally classified into operator ensembles, algorithm ensembles, and hybrid deterministic–stochastic ensembles. Operator ensembles employ multiple variation operators within a single framework. For example, ensemble Differential Evolution (DE) \cite{mallipeddi2011differential,li2013adaptive,sallam2017landscape} may combine “DE/rand/1,” “DE/current-to-best/1,” and “DE/best/2” strategies with adaptive operator selection, while ensemble PSO variants \cite{lynn2017ensemble,akkaya2025psopm} may use multiple inertia or update rules across particles.

An emerging instance of large-scale algorithm ensembles is Animorphic Ensemble Optimization (AEO), which frames a flexible, large-scale island model combining multiple evolutionary and swarm algorithms \cite{price2023animorphic}. In AEO, each island maintains a distinct optimizer (e.g. DE, PSO, ES, GWO, WOA, etc.), and the system proceeds in repeated cycles: within each cycle, each optimizer runs for a given number of generations, then migration or redistribution of population members occurs based on performance metrics. AEO formalizes mobility metrics to decide how many individuals migrate and in which direction, adjusting dynamically over time to emphasize more successful algorithm islands. Experimental results in the original paper attest to AEO’s effectiveness on benchmark problems: by leveraging heterogeneous optimizers in a large-scale island ensemble, AEO tends to outperform individual constituent algorithms and demonstrates resilience against the NFL pitfalls. This example clearly shows how ensemble models can outperform the singular optimization algorithms. 

Engineering optimization problems frequently involve multiple objectives that must be addressed simultaneously. In the energy domain, these objectives span economic, socioeconomic, efficiency, and safety considerations. Taking nuclear power as an example, due to the application focus of this paper, these conflicting objectives arise in applications such as core reload pattern optimization \cite{seurin2024physics, seurin2024multi, shaukat2021core}, reactor siting and licensing trade-offs \cite{erdem2025multi}, and microreactor control and fuel usage optimization \cite{price2022multiobjective,tunkle2025nuclear}. Such problems require multiobjective optimization (MOO) algorithms capable of capturing trade-offs and producing Pareto fronts \cite{ehrgott2002multiple} by identifying non-dominated solutions. A solution $\mathbf{x}_1$ dominates $\mathbf{x}_2$ if it is no worse in all objectives and strictly better in at least one. Multiobjective algorithms, therefore, aim for both convergence toward the true Pareto-optimal front and adequate diversity across it. A central mechanism for achieving this is \emph{non-dominated sorting}, used in NSGA-II and NSGA-III \cite{deb2002fast,deb2013evolutionary}. In the single-objective case with a scalar-valued fitness function $f$, comparison reduces to:

\begin{equation}
    \min_{\mathbf{x} \in \Omega} f(\mathbf{x})
    \;\;\Rightarrow\;\;
    f(\mathbf{x}_2) < f(\mathbf{x}_1)
    \;\;\Rightarrow\;\;
    \mathbf{x}_2 \;\text{dominates}\; \mathbf{x}_1,
    \label{eq:so_dominance}
\end{equation}

\begin{equation}
    \mathbf{x}^* = \arg\min_{\mathbf{x} \in \Omega} f(\mathbf{x})
    \;\;\Rightarrow\;\;
    \mathbf{x}^* \;\text{dominates all}.
    \label{eq:global_min}
\end{equation}

In the multiobjective case, dominance is generalized to $n$ objectives:

\begin{equation}
    f_i(\mathbf{x}_2) \leq f_i(\mathbf{x}_1), \;\forall i \in \{1,\dots,n\}, \;\;
    \exists j : f_j(\mathbf{x}_2) < f_j(\mathbf{x}_1)
    \;\;\Rightarrow\;\;
    \mathbf{x}_2 \;\text{dominates}\; \mathbf{x}_1.
    \label{eq:ndsort}
\end{equation}

Widely used MOO algorithms adapt single-objective frameworks to multiobjective settings by incorporating mechanisms for dominance-based selection, diversity preservation, and structured search guidance. These methods can be grouped into several complementary families: dominance-based methods such as NSGA-II \cite{deb2002fast} and SPEA2 \cite{zitzler2001spea2}, which weight individuals by dominance relations; reference-vector approaches such as RVEA and NSGA-III \cite{cheng2016reference,deb2019reference}, which promote uniform coverage through predefined directions in objective space; decomposition-based algorithms such as MOEA/D \cite{zhang2007moead}, which solve scalar subproblems using weighted-sum or Tchebycheff formulations; and indicator-based methods such as SMS-EMOA \cite{beume2007sms}, which rely on hypervolume (HV) contributions despite their computational cost in higher dimensions. Together, these approaches provide complementary treatments of convergence, diversity, and objective-space structure and represent the core algorithmic families in contemporary multiobjective optimization.

Nuclear power is a domain where designs must satisfy a diverse set of often conflicting goals: multiple safety criteria, economic performance measures, and manufacturing or regulatory constraints. Because of these competing demands, many researchers have applied optimization, machine learning, and artificial intelligence methods to multiobjective problems in nuclear engineering, such as reactor core design, power plant layout, fuel cycle planning, and control systems. In earlier work, optimization algorithms have been used to solve core loading problems \cite{zhao1998fuelgen, seurin2024physics, seurin2024multi, seurin2025impact, zhang1995optimization}, fuel management and enrichment allocation \cite{dechaine1996fuel}, and plant-level layout and cost minimization problems \cite{pereira2003coarse}. Other studies have focused on applying multiobjective evolutionary algorithms to core design and safety trade-offs \cite{yang1999application, zhang2015multiobjective}, integrated power plant design \cite{wilding2020use}, and criticality search for small modular reactor (SMR) designs \cite{gu2023openneomc}. Machine learning and artificial intelligence (AI) approaches have also been adopted to accelerate the assessment of nuclear systems and facilities, particularly through surrogate modeling of nuclear accidents \cite{radaideh2020neural}, reactor physics \cite{radaideh2020surrogate}, fault detection in particle accelerators \cite{radaideh2023early}, and safety margins in nuclear power plants \cite{seurin2025impact, che2022machine}. These methods reduce the computational cost of design iterations by approximating expensive reactor simulations. The increasing research volume in this area demonstrates that optimization and AI are rapidly becoming indispensable tools for nuclear design. By definition, nuclear reactor simulations are computationally expensive. To solve these problems effectively, more reliable, cost-effective, and competent MOO algorithms are required. This has led to the development of hybrid frameworks, surrogate-assisted approaches, and ensemble strategies that balance accuracy with efficiency in high-stakes nuclear power applications.

\subsection{Novelty}

This research investigates an ensemble-based MOO algorithm, examining its performance on mathematical benchmark functions relative to other MOO methods through systematic evaluation of algorithmic performance across multiple problems, and demonstrating its applicability to computationally expensive engineering applications involving nuclear reactor design optimization. While similar ensemble approaches have been studied for single-objective problems, their extension to MOO has not been explored in the existing literature. 

The Multiobjective Animorphic Ensemble Optimization (MAEO) method uses the foundation created by the AEO algorithm \cite{price2023animorphic}. We extend the AEO framework to handle MOO, resolving the challenges introduced by Pareto ranking, diversity management, and Pareto front coverage. The created ensemble-based MOO method is tested on mathematical benchmark functions. In this work, the analysis of the MAEO algorithm is presented and its performance is proven. Furthermore, the MAEO algorithm has been employed in the first-cycle fuel enrichment optimization of an SMR design. Even though the exact details of the SMR core designs are trade secrets and not released, academic research papers tend to use approximations of the SMR core designs to study their neutronic and thermal-hydraulic characteristics. To test the MAEO algorithm and run first cycle assembly enrichment optimization on a nuclear reactor core, the NuScale-like core description has been selected \cite{fridman2023nuscale}. Deterministic reactor core simulation codes CASMO4 and SIMULATE3 are used to evaluate different designs with low computation cost and high accuracy. Economic and safety objectives of the core designs have been evaluated in every generation. For every simulation, the levelized cost of electricity (LCOE), fuel cycle length expressed in fuel effective full power years (fuel EFPY), maximum boron concentration in the cycle, power peaking factor (F$_q$) and enthalpy-rise hot channel factor (F$\Delta$H) safety parameters have been extracted. These parameters are used for evaluating the performance of the investigated core designs. 

Our work provides a novel ensemble-based MOO algorithm to the optimization field, describing the underlying design choices and its complete methodology, showing its performance and reliability, proving its effectiveness in multiple benchmark functions and on a real-world optimization problem. Accordingly, the key contributions and findings of this study can be summarized as follows:  

\begin{enumerate}

\item Development of the \textbf{Multiobjective Animorphic Ensemble Optimization (MAEO)} framework, enabling parallel multiprocessing and coordinated multi-algorithm search for high-cost multiobjective optimization.

\item Establishment of a \textbf{hypervolume-based performance criterion to compare the ensemble algorithms}, identifying hypervolume as the most robust and parameter-free indicator for comparing multiobjective algorithms in black-box settings.

\item Introduction of a \textbf{Pareto individual scoring formulation} that enforces strict Pareto-rank separation and augments selection through integrated crowding-distance and nadir-distance measures.

\item Application of MAEO to the \textbf{equilibrium-cycle optimization of a small modular reactor core}, demonstrating its effectiveness for discrete, constrained, and simulation-intensive reactor design problems.

\end{enumerate}

The remaining sections are structured as follows: Section \ref{sec:MAEO} introduces the proposed Multiobjective Animorphic Ensemble Optimization (MAEO) framework, outlining its algorithmic components and key mechanisms for migration, coordination, and convergence. Section \ref{sec:opt_verif} presents the performance evaluation of MAEO, including a single-problem benchmark and a comprehensive Wilcoxon signed-rank statistical analysis across established multiobjective test functions. Section \ref{sec:nuclear} presents an engineering application of MAEO to equilibrium-cycle core optimization of an SMR-like reactor, and an assessment of multiprocessing efficiency. Finally, Section \ref{sec:conclusion} summarizes the main findings and discusses future research directions.

\section{Multiobjective Animorphic Ensemble Optimization}
\label{sec:MAEO}

To extend single-objective ensemble algorithms to MOO problems, the functioning of the MAEO algorithm is described in the following subsections. Its methodology is compared with AEO, its migration phase is examined, and its performance is benchmarked to highlight the algorithm’s capabilities.

\subsection{Algorithm Description}

\begin{figure}
\centering
\includegraphics[width=\linewidth]{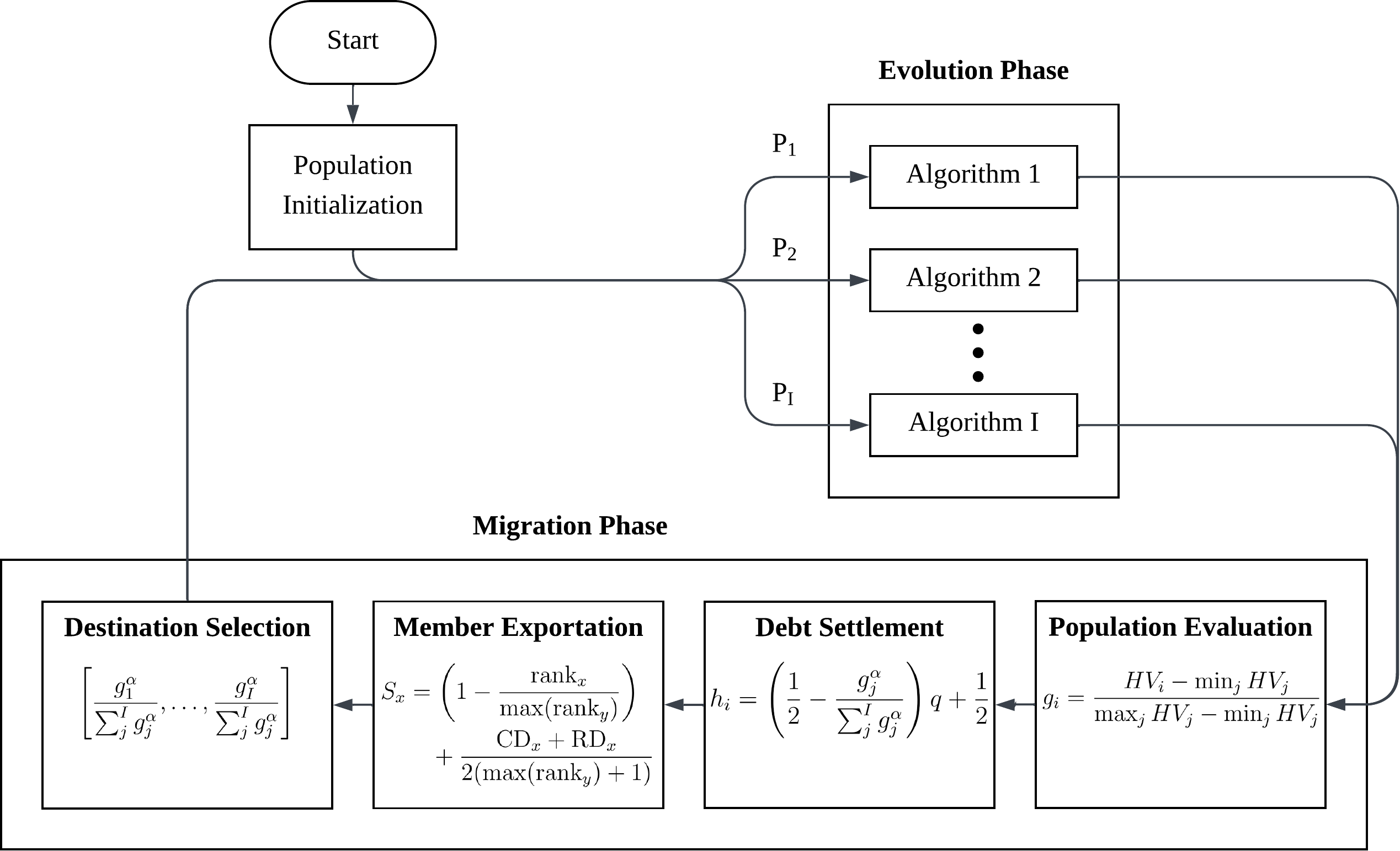}
\caption{Flow diagram of the MAEO algorithm.}
\label{fig:maeo_phases}
\end{figure}

The MAEO method consists of two phases executed sequentially: the \textit{evolution phase} and the \textit{migration phase}. In MAEO, each MOO algorithm (e.g., NSGA-III \cite{deb2013evolutionary}, SPEA2 \cite{zitzler2001spea2}) is defined as an island. During the evolution phase, these islands independently evolve and return their populations. The input and output individuals of the final generation from this phase are then passed to the migration phase. The migration phase itself is composed of four steps: population evaluation, debt settlement, member exportation, and destination selection. The flow diagram of the MAEO algorithm is given in Figure~\ref{fig:maeo_phases}. The flow of the migration phase of the MAEO algorithm is described in the following subsections.

\subsubsection{Population Evaluation}
\label{sec:pop_eval}

The first step in the migration phase is to evaluate the relative performance of the optimization algorithms (islands) and calculate the island debts. The debt refers to the number of individuals an island owes to other islands. The island performance is measured by the HV indicator. The HV indicator can be formally defined as the Lebesgue measure of the portion of the objective space that is weakly dominated by the Pareto set $P$ and bounded by a reference point $\mathbf{r}$ \cite{ZitzlerThiele1999}. As an \textbf{indicator}, it assigns a single scalar value to the entire Pareto set, allowing different approximation sets to be compared quantitatively based on their performance. The Pareto set $P$ is constructed by evaluating the fitness values of the decision vectors $\mathbf{x}\in S$ and performing non-dominated sorting on the resulting objective vectors. The resulting set contains $N$ Pareto-optimal points, each denoted by $\mathbf{p}\in P$. The mapping from decision vectors to their objective vectors is given by:

\begin{equation}
\mathbf{f} = (f_1,\dots,f_n), \qquad
\mathbf{p} = \mathbf{f}(\mathbf{x}) = [f_1(\mathbf{x}),\dots,f_n(\mathbf{x})] = (p_1,\dots,p_n),
\qquad
\mathbf{p} \in P = \mathbf{f}(S).
\end{equation}
where $\mathbf{f}$ is the $n$-dimensional vector–valued objective function, composed of $n$ distinct scalar objective functions $f_1,\dots,f_n$, each mapping the decision space to a real-valued performance measure. 

The reference point represents a worst-case objective vector. In practice, it is often taken as the nadir point or a slightly larger multiple of it to ensure all solutions lie within the dominated region. To calculate the HV of a region, for each objective vector $\mathbf{p} = (p_1, \dots , p_n)$, an axis-aligned hyperrectangle is constructed, spanning from $\mathbf{p}$ to $\mathbf{r}$. The HV is then the measure of the union of all such hyperrectangles, which avoids double counting of overlapping dominated regions. This general form of HV is given in Eq. \eqref{eq:hvol}.

\begin{equation}
    \text{HV(P)} = \text{Leb}\left( \bigcup_{\mathbf{p} \in P} [p_1, r_1] \times \cdots \times [p_n, r_n] \right),
    \label{eq:hvol}
\end{equation}

The operator $\text{Leb}(\cdot)$ denotes the \emph{Lebesgue measure} (generalization of length, area, and volume) applied to the union of hyperrectangles spanning from each $\mathbf{p}$ to the reference point $\mathbf{r}$. In other words, it calculates the n-dimensional volume of the dominated region by constructing hyperrectangles from each non-dominated objective vector. 

The calculation of HV can be nontrivial. The calculation begins with the one-dimensional case. In the one-dimensional case, \textbf{p}=p$_1$ and \textbf{r}=r$_1$. The $\text{HV}^{(1)}$ is simply defined as the length of the interval between the best Pareto point and the reference point, which corresponds to the union of all intervals defined by the Pareto set, as shown in Eq.~\eqref{eq:hv_1d}:

\begin{equation}
    \text{HV}^{(1)}(P, r_1) = \max \bigl(0,\, r_1 - \min_{p_1 \in P} p_1 \bigr).
    \label{eq:hv_1d}
\end{equation}

For higher dimensions, the computation is recursive. The Pareto points are first sorted with respect to the last objective dimension, producing an ordered sequence of “slices” along that axis. For each slice, the thickness is calculated as the difference between consecutive coordinate values in the last dimension (or between the reference point and the last Pareto point for the boundary slice). Within each slice, the set of Pareto points that remains valid (i.e., those not dominated in earlier dimensions) is projected into a lower-dimensional space by discarding the last coordinate. The HV contribution of the slice is then computed as the product of its thickness and the HV of the lower-dimensional projection. Summing the contributions of all slices yields the total, n-dimensional $\text{HV}^{(n)}$. \emph{The recursion starts at d = n and terminates at d = 1}. This recursive formulation is given in Eq.~\eqref{eq:hv_nd}:

\begin{equation}
    \text{HV}^{(d)}(P, \mathbf{r}) = \sum_{k=1}^{L-1} \bigl( z_{k+1} - z_{k} \bigr)\,\text{HV}^{(d-1)}\!\bigl(P_k,\, \mathbf{r}_{1:(d-1)}\bigr),
    \label{eq:hv_nd}
\end{equation}
where $z_1 < z_2 < \cdots < z_L = r_d$ are the distinct scalar values of the $d$-th objective extracted from all points in $P$ and augmented with the reference coordinate $r_d$, and $\,\mathbf{r}_{1:(d-1)} = (r_1,\dots,r_{d-1})^\mathsf{T}$ denotes the projection of the reference point onto the first $d$–$1$ objectives. Here $L$ denotes the total number of distinct slice boundaries obtained from the sorted scalar values of the last objective after including the reference coordinate; it is therefore determined by the data and the reference point rather than by the problem dimension. Thus each $z_k$ is a real number, not a $d$-dimensional point. For $k = 1,\dots,L-1$ we define:

\begin{equation}
    P_k = \bigl\{ \mathbf{p}\in P : p_d \le z_k \bigr\}.
\end{equation}
which is the subset of points whose $d$-th objective value does not exceed the lower boundary $z_k$ of the slice $(z_k, z_{k+1}]$. These are precisely the points that dominate the entire slab in the $d$-th objective. To apply the recursive hypervolume formula, we slice along the $d$-th objective as follows:

\begin{enumerate}
    \item Extract the $d$-th objective values $\{p_d : \mathbf{p}\in P\}$, sort them in ascending order, and append the reference coordinate $r_d$. This yields scalar breakpoints $z_1 < z_2 < \cdots < z_L = r_d$, where each $z_k$ is a real number (not a $d$-dimensional point).
    
    \item For each interval $(z_k, z_{k+1}]$, determine the set of points that dominate this entire slab in the last objective, namely $P_k = \{\mathbf{p}\in P : p_d \le z_k\}$. These points contribute to the hypervolume of that slice through their projection onto the first~$d-1$ objectives.
    
    \item The $d$-dimensional hypervolume is then obtained by summing the contributions of all slabs: each slab has width $(z_{k+1} - z_k)$ in the $d$-th dimension and height $\text{HV}^{(d-1)}(P_k, \mathbf{r}_{1:(d-1)})$ from the $(d-1)$-dimensional hypervolume of the dominating subset.
\end{enumerate}

\begin{figure}
\centering
\includegraphics[width=0.35\linewidth]{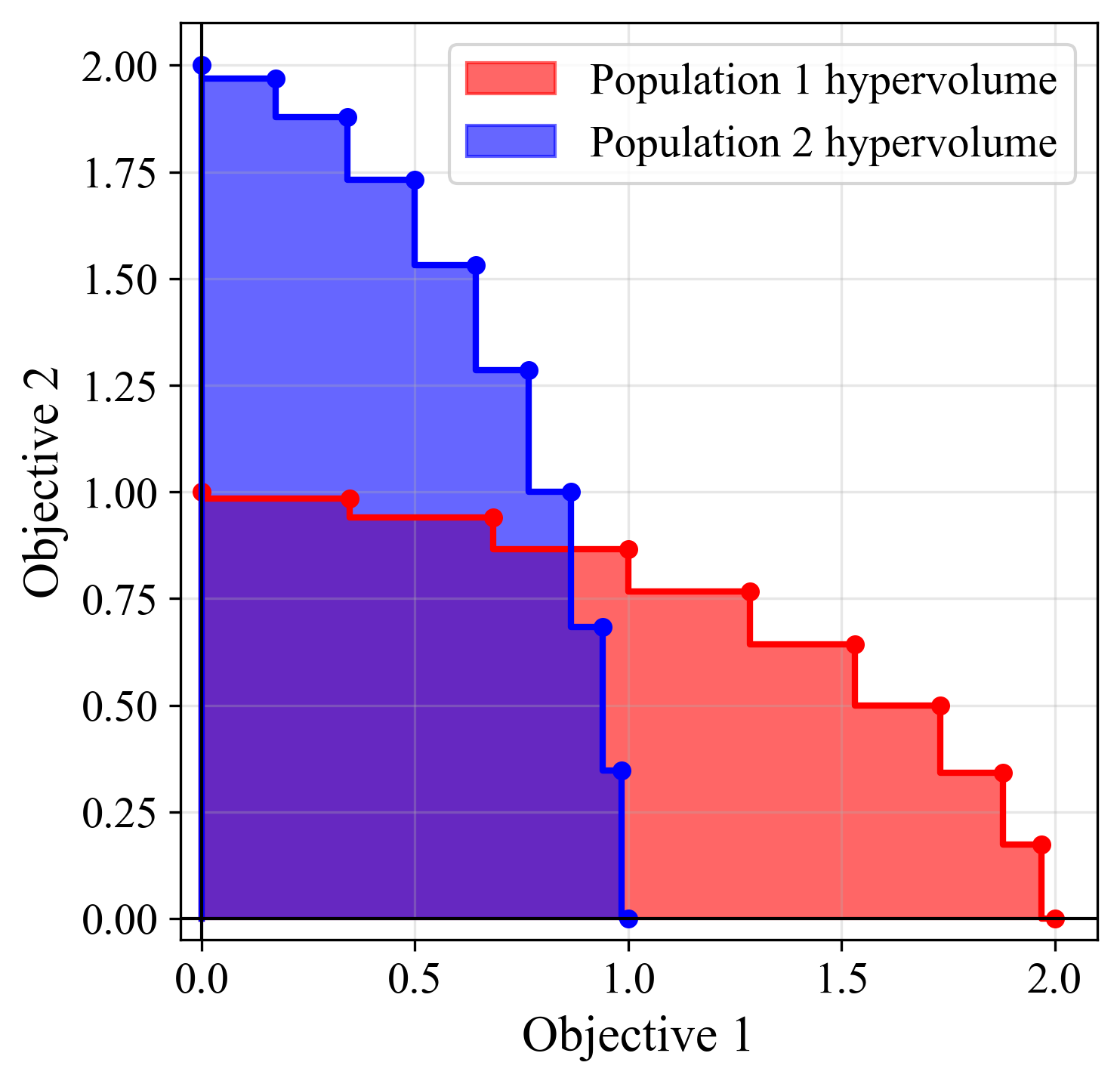}
\caption{Demonstration of hypervolume computation for optimization islands, where each island’s HV is calculated as the dominated volume bounded by its Pareto front and the reference box.}
\label{fig:hypervol_demo}
\end{figure}

The concept of hypervolume in two dimensions is illustrated in
Figure~\ref{fig:hypervol_demo}. HV calculation is computationally demanding, especially for large Pareto fronts in high dimensions. The exact recursive calculation has a worst-case complexity of $O(N^{n-2}\log(N))$, which grows rapidly with the number of individuals $N$ and objectives $d$. To manage this, an empirical threshold is applied: when $N^{n-2}\log(N) > 2\times 10^{6}$, where the single-core runtime exceeds 30 seconds, the algorithm switches from exact to approximate computation.  

Approximation is performed via Monte Carlo sampling within a bounding box defined by the front’s maximum objective values and the reference point. Batch sizes are chosen as $5000 \times \max(1,\log(\text{Box Volume}))$, clamped between 5000 and 50,000. The box volume is defined as the volume of the box that encompasses the reference point and the minimum objective values in the computation. Iterations continue until the relative error between successive estimates falls below 0.1\%, ensuring sufficient accuracy.  

We define $\text{HV}_i(c)$ as the hypervolume indicator computed for MAEO island $i$ at cycle $c$. This value represents the dominated volume contributed by the island's current Pareto set in that iteration of the optimization.

In reference to alternative MOO metrics, the HV indicator result [$\text{HV}_i(c)$] is retained as the performance metric in place of inverse generational distance (IGD) or the $\epsilon$-indicator. IGD requires a predefined reference Pareto front, which may be unavailable or inferior to the obtained solutions, and it primarily reflects convergence while only weakly representing diversity. The $\epsilon$-indicator is even more restrictive, measuring convergence exclusively. In contrast, HV simultaneously captures both convergence and diversity without relying on external references. Approximate HV methods such as HypE and related Monte Carlo schemes are widely adopted in multiobjective evolutionary algorithms ~\cite{bringmann2013parameterized}. Despite its computational cost in higher dimensions, these advantages justify its continued use, supplemented by approximation techniques when necessary.

At the end of each evolution phase, the MAEO algorithm computes the HV for every island. These HV values serve as the basis for calculating each island’s performance score. In the original single-objective AEO algorithm \cite{price2022aeo}, island performance was measured by the change in fitness between successive cycles. In single-objective optimization, the fitness of an elitist algorithm \cite{deb2002fast} is absolute and changes only when a better solution is found. However, in multiobjective problems, island HV values fluctuate slightly across cycles as new points are explored, introducing small variations in the computed HV values.

These fluctuations can overly influence migration decisions during late cycles. While AEO evaluates island performance using the change in fitness, applying this principle directly to HV in multiobjective settings leads to unstable migration behavior. To improve stability, the MAEO framework uses the absolute HV indicator rather than its change between cycles.

Using the HV indicator as the island fitness measure offers several advantages: it is parameter-free, scales naturally with the number of objectives, and consistently rewards high-performing islands in later stages of optimization, thereby promoting convergence. Several alternative HV-based island–performance formulations were also tested, including (i) a linear HV weighting of the form $A(\text{HV}_i(c-1)) \cdot \text{HV}_i(c)$, where $A$ is a linear weight that increases with the previous-cycle HV value ($\text{HV}_i(c-1)$) to reward consistently strong islands; (ii) HV improvements from the last cycle $\Delta \text{HV}_i(c) = \text{HV}_i(c) - \text{HV}_i(c-1)$; (iii) Gaussian-normalized HV improvements; and (iv) raw HV values $\text{HV}_i(c)$. The first three variants led to convergence instabilities and poor performance, whereas raw HV values proved stable and were therefore adopted as the preferred formulation. Consequently, the HV indicator is adopted as the sole parameter-free performance metric for black-box optimization problems where the ideal Pareto front is unknown.

\subsubsection{Debt Settlement}
\label{sec:debt}

For notational simplicity, all quantities from this point onward refer to the current MAEO cycle $c$. For brevity, the cycle index $(c)$ is omitted from $\text{HV}_i(c)$ and all other cycle-dependent terms.

After computing island fitness values, these values are used for calculating the number of individuals each island will export and receive. The MAEO framework applies min–max scaling to normalize island HV indicators into the range $[0,1]$, yielding the scaled indicator $g_i$ in Eq. \eqref{eq:gindicator}.
\begin{equation}
g_i = \frac{\text{HV}_i - \min_j \text{HV}_j}{\max_j \text{HV}_j - \min_j \text{HV}_j}
\label{eq:gindicator}
\end{equation}

The normalized performance $g_i$ determines the number of individuals to remove from each island, denoted $e_i$, which is sampled from a binomial distribution as follows

\begin{equation}
e_i \sim \text{Binomial}(N_i, h_i),
\label{eq:binomial}
\end{equation}
where $N_i$ is the island population size. The success probability $h_i$ of this binomial sampling depends on both the normalized island performances and the current cycle number, as defined in Eq.~\eqref{eq:h_i}:

\begin{equation}
h_i = \left(\frac{1}{2} - \frac{g_i^{\alpha}}{\sum_j^I g_j^{\alpha}}\right)q + \frac{1}{2},
\label{eq:h_i}
\end{equation}
where $q$ and $\alpha$ are cycle-dependent parameters evolving according to Eq.~\eqref{eq:q} and~\eqref{eq:alpha}:

\begin{equation}
q = \frac{2(1-c)}{1-N_{cyc}} - 1,
\label{eq:q}
\end{equation}

\begin{equation}
\alpha = \frac{c - 1}{N_{cyc} - 1},
\label{eq:alpha}
\end{equation}
with $N_{cyc}$ denoting the total number of cycles. As the number of current cycle $c$ increases, $q$ transitions from $-1$ to $1$, and $\alpha$ increases from $0$ to $1$. This design ensures that migration remains gradual during early cycles—encouraging exploration—and becomes increasingly aggressive in later cycles, promoting exploitation and convergence toward the best-performing islands.

Initially, this formulation of $h_i$ distributes migration almost evenly across islands, allowing each to explore the fitness landscape. As the number of cycles grows, $q$ increases linearly, enhancing the effect of accumulated island “debt” on the number of exported individuals. Concurrently, rising $\alpha$ reduces the import rate of individuals to well-performing islands while nonlinearly increasing exports from poorly performing ones. Through this coupled mechanism, all islands are afforded sufficient opportunity to explore during early stages, whereas in later cycles, weaker islands are gradually eliminated and stronger ones attract the majority of available computational resources.

\subsubsection{Member Exportation}
\label{sec:export}

After determining $e_i$, the next step is to select which individuals to export. In single-objective optimization, the selection is straightforward, typically based on the fitness value. In MOO, however, an individual’s quality must consider both convergence and diversity.  

To this end, MAEO defines an individual performance score $S_x$, which combines Pareto rank, crowding distance, and reference distance, as shown in Eq.~\eqref{eq:indiv_score}:

\begin{equation}
    S_x \;=\; 
    \left( 1 - \frac{\text{rank}_x}{\max(\text{rank}_y)} \right)
    \;+\;
    \frac{\;\text{CD}_x + \text{RD}_x\;\;}{2(\max(\text{rank}_y) + 1)},
    \label{eq:indiv_score}
\end{equation}
where $y \in [1, ..., N_i]$ is the index of individuals in island $i$, $\text{rank}_x$ is the Pareto rank of input vector $x$ (lower is better), $\text{CD}_x \in [0, 1]$ is the normalized crowding distance, and $\text{RD}_x \in [0, 1]$ is the min-max normalized reference distance from the nadir point. 

The two terms in Eq.~\eqref{eq:indiv_score} benefit from clarification. The reference distance (RD) is computed as the Euclidean distance between an individual and the nadir point in the $m$-dimensional objective space, $RD_x = \| \mathbf{f}_x - \mathbf{r} \|_2$, and is then min–max normalized using the minimum and maximum distances observed within the same population. The crowding distance (CD) is defined separately for each Pareto rank: for each objective, individuals in the same rank are sorted by objective value, and the normalized spacing between each individual and its immediate neighbors is computed. The per-objective contributions are summed to yield the CD of each individual, after which all CD values in the rank are normalized to $[0,1]$. Following standard practice, boundary individuals in every objective are assigned a CD value of 1 to promote exploration of the edges of the Pareto front. This procedure is repeated independently for every rank. Together, CD and RD quantify local spread and global positioning, allowing the scoring function to reward both within-front diversity and movement toward the outer boundary of the Pareto front.

\begin{figure}
\centering
\includegraphics[width=0.3\linewidth]{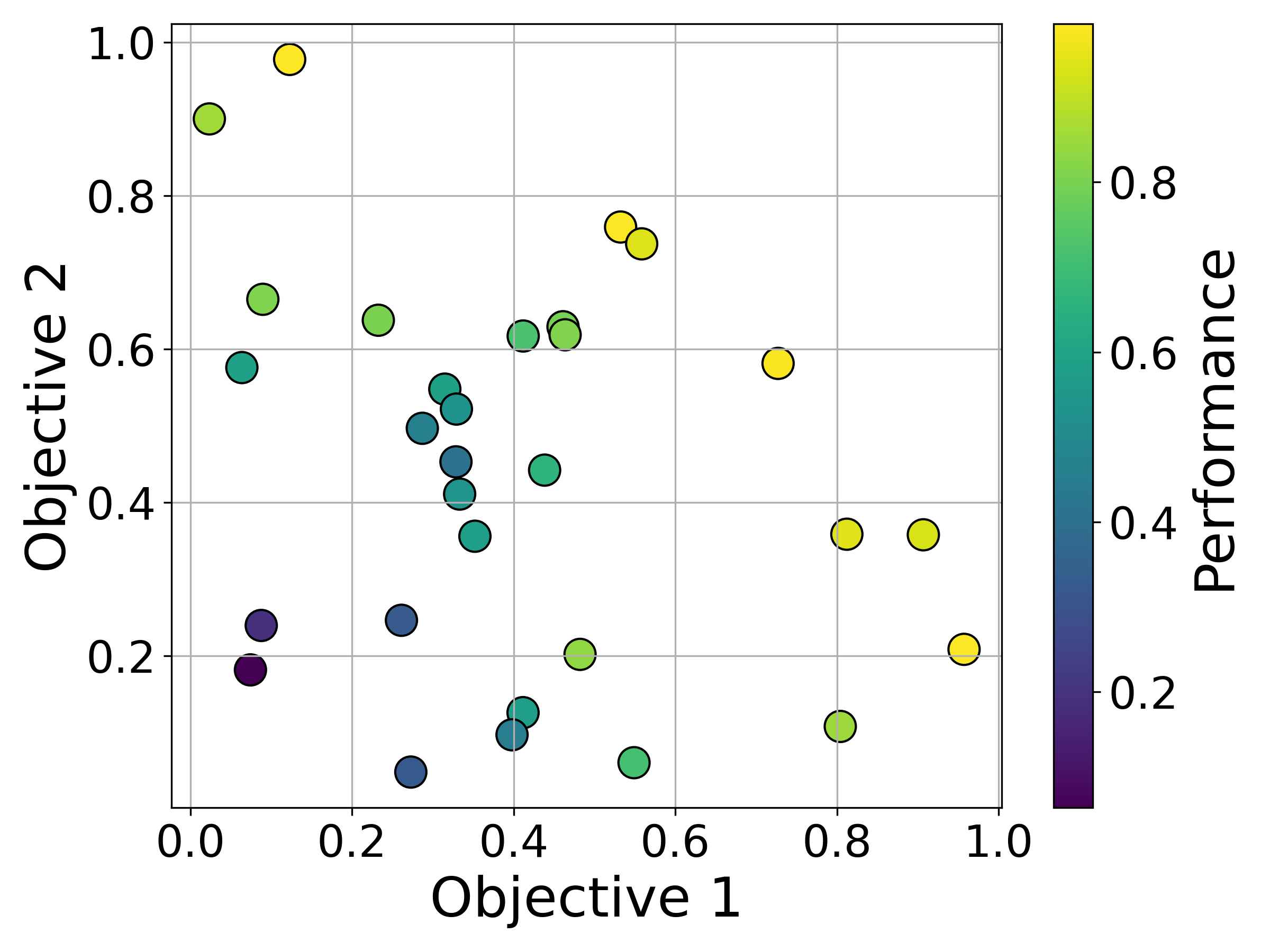}
\caption{Demonstration of the individual performance score.}
\label{fig:ind_per_demo}
\end{figure}

This score has two roles. The rank component enforces convergence by prioritizing non-dominated individuals. The diversity component balances exploration by promoting solutions that are both farther from the nadir point ($RD_x$) and better spaced along the front ($CD_x$). The denominator of the second term is the band scaling factor. By using this factor 
\Big($\frac{1}{2(max(rank_y)+1)}$\Big), \textit{strict rank separation} is guaranteed: no dominated solution can ever achieve a higher score than any non-dominated solution, regardless of its diversity values. This ensures that convergence pressure toward the Pareto front is preserved, while diversity still influences selection within the same rank. The effect of the individual performance score is illustrated in Figure~\ref{fig:ind_per_demo}, where it is applied to randomly generated points in the first quadrant of the unit circle.

\subsubsection{Destination Selection}

Finally, the $e_i$ worst-performing individuals selected for removal are pooled together across all islands. Let $K$ denote the total number of individuals in this pool. These individuals are then redistributed among the islands using a multinomial distribution, where the probability of receiving individuals is proportional to the normalized HV of the island. The allocation rule is defined in Eq.~\eqref{eq:multinomial}:

\begin{equation}
    \rho_i \in \bm{\rho} \quad,\quad \bm\rho \sim \text{Multinomial} \left( K, \left[\frac{g_1^\alpha}{\sum_j^I g_j^\alpha}, \dots ,\frac{g_I^\alpha}{\sum_j^I g_j^\alpha} \right]\right),
    \label{eq:multinomial}
\end{equation}
where $\boldsymbol{\rho}$ denotes the vector of incoming individuals, with $\rho_i$ representing the number assigned to each island $i$, and $I$ the total number of islands. This formulation ensures that islands with higher performance (lower debt) receive more individuals and thus more computational resources, reinforcing their success.  

Once $\bm{\rho}$ is sampled, the $K$ individuals are randomly distributed among the islands. Within each island, incoming individuals are treated equally; their performance scores do not affect their placement. 

If an island has no remaining individuals after the migration step, the island is disabled, and any individuals it previously held are rescued and distributed following the same destination selection method described above. The disabled island is not considered for performance scoring. The island disabling and rescued individual redistribution step completes the migration phase.

After the migration phase, the next optimization cycle begins, where the algorithms are executed for the predefined number of generations. The reference point is determined only after these runs are completed. This ensures that HV calculations are consistently based on the global nadir point, thereby preventing error propagation across generations. However, this approach can cause a slight reduction in the maximum observed HV between cycles. This reduction serves to correct the error introduced by relying on an approximate nadir point during the first cycle of the optimization process.

\subsubsection{Convergence Assist}
\label{sec:conv_assist}

The MAEO framework integrates multiple optimization algorithms operating as islands. During the initial cycles, the algorithm allocates more resources to weaker-performing islands to enhance exploration. In later cycles, however, resources are increasingly redirected toward stronger islands to promote convergence. In problems where a single multiobjective algorithm is inherently more effective, allocating effort to multiple algorithms may slow convergence. To mitigate this, a new feature called \textbf{convergence assist} is introduced.

When convergence assist is enabled, the number of cycles in the migration phase is reduced by a specified ratio, denoted as $CA$. The computational resources saved from this reduction are then devoted to additional optimization using the best-performing island identified at the end of the migration phase. For instance, if the total number of cycles is defined as $C=25$ and the convergence assist ratio as $CA=0.2$, the MAEO runs its evolution and migration phases for 20 cycles, while the final 5 cycles are executed solely on the best-performing algorithm. The populations from all remaining islands are merged into that island, and subsequent cycles proceed without further migration.

Conducting the final optimization cycles using a single algorithm ensures that the resulting solutions remain sufficiently diverse to form a well-distributed Pareto front. \textbf{This requirement distinguishes MOO from single-objective problems}. In single-objective optimization, convergence to a single global optimum is the primary goal. A single individual with best fitness value is enough to represent the best solution the island or ensemble algorithm finds. However, in case of MAEO, a large population is required to represent the island and present the final Pareto front of the ensemble algorithm. The implementation of convergence assistance is rooted in this fundamental difference. It accelerates convergence in the final stages of optimization while preserving solution diversity.

Empirical results show that MAEO naturally merges most islands into a single dominant island by approximately 95\% of the total cycle count. The convergence-assist mechanism formalizes and accelerates this process, enabling faster consolidation of resources and allowing additional refinement of the final Pareto front through extended optimization on the most successful algorithm.

\subsection{Migration Phase Diagnostics}
\label{sec:migration}

In an ensemble optimization algorithm, it is important to show the relation of the individual algorithms. In the case of MAEO, this interaction happens through migration. To keep track of the migration speed in the algorithm, the migration of an individual is defined as a single step in a Markov chain in this section. The mobility matrix is defined in Eq.\eqref{eq:mobility} to describe the movement of individuals across the populations.

\begin{equation}
    M_{j,i} = Pr(\mathbf{x} \in \Pi_i \rightarrow \Pi_j).
    \label{eq:mobility}
\end{equation}

As directly adapted from the AEO algorithm, the event of $\mathbf{x} \in \Pi_i \rightarrow \Pi_j$ is defined as the event where a member selected for exportation in island $i$ at the beginning of the migration phase ends up in island $j$. The probability of an individual being exported from population $i$ is assumed uniform, as all individuals have equal opportunity to be the best and the worst in the population in each cycle. This simplified individual selection probability is given by $w_k = \tfrac{1}{N_i}$, where $N_i$ is the number of individuals in island $i$.

Using this notation allows the island parameters $g_i$ and $h_i$ to be represented within a square matrix $M$, which captures the mobility of exported individuals between islands. The properties of the matrix $M$ define the migration probabilities during a single cycle. Based on the probability mass functions of the migration equations given in Eq.~\eqref{eq:binomial} and~\eqref{eq:multinomial}, the migration probability $M_{i,j}$ can be expressed as shown in Eq.~\eqref{eq:Mij_calc}. A detailed derivation of $M_{i,j}$ is available in the original AEO formulation by Price et al.~\cite{price2023animorphic}.

\begin{equation}
    M_{j,i} \;=\;
    \begin{cases}
            \delta_{i,j}+\left( \frac{g_j^\alpha}{\sum g_k^\alpha} - \delta_{i,j} \right) \; \sum_{k=0}^{N_i} \binom{N_i}{k} (h_i)^k(1-h_i)^{N_i-k}, & \text{if } N_i > 0,\\[8pt]
        \frac{1}{I}, & \text{if } N_i = 0,\\[8pt]
    \end{cases}
    \label{eq:Mij_calc}
\end{equation}
where $\delta_{i,j}$ is the Kronecker delta function.

The created square matrix $M$ represents migration probabilities by considering the migration as a one-step Markov chain. This non-negative matrix has its largest eigenvalue equal to $1$, and the eigenvector corresponding to this eigenvalue is the Perron–Frobenius eigenvector, i.e., the stationary distribution. The second largest eigenvalue in magnitude, $\lambda_2$, controls the mixing speed: it determines how fast the chain converges to its stationary distribution.

In this work, we use the \textit{heuristic speed indicator} of the short-term rate of state changes described in \cite{price2023animorphic}, defined as the reciprocal of $\lambda_2$. We denote this speed indicator by $\theta$:
\begin{equation}
    \theta = \frac{1}{\lambda_2}.
\end{equation}

A second indicator is the \textit{average mobility index}, widely used in demography, economics, and some engineering applications of Markov chains. Here, $\mathrm{tr}(M)$ is the sum of the diagonal entries, i.e., the probability of remaining in the same state. $\mathrm{tr}(M)=0$ indicates maximum mobility, while $\mathrm{tr}(M)=I$ implies the chain is completely immobile. Based on this, the normalized trace-based mobility index, $\phi$, is defined as
\begin{equation}
    \phi = \frac{I-\mathrm{tr}(M)}{I-1}.
\end{equation}

It is important to characterize how rapidly the states of the system evolve during the migration phase. The Perron–Frobenius eigenvector describes the stationary distribution toward which the chain converges, while the eigenvalues provide information on the speed of this convergence. In particular, the heuristic indicator $\theta$ quantifies the short-term dynamics, whereas the normalized trace-based mobility index $\phi$ reflects the average long-term tendency of the system to remain in or leave its current state. Together, these indicators provide complementary perspectives on the convergence properties of the migration process. To illustrate the evolution of island populations with respect to these indicators, the population distributions are presented as a function of cycle number when the results are presented later.

\subsection{Parallelism Capability}
MAEO is designed with a nested multiprocessing architecture that the original AEO did not support \cite{price2023animorphic}. MAEO enables parallelism at two hierarchical levels. At the lower level, each island can evaluate its individuals using multiple worker processes, up to the limit imposed by its population size. At the upper level, multiple islands can execute concurrently, allowing full parallel operation across the island model. For instance, four islands running simultaneously with 25 lower-level workers each produce 25 $\times$ 4 = 100 active fitness-evaluation processes, in addition to a small number of controller and coordination processes. This structure allows MAEO to exploit large compute nodes and multi-node clusters by distributing work both within and across islands.

This nested design was validated using both the engineering application presented in this paper (e.g., the nuclear SMR optimization) and synthetic delay–based test functions. These complementary setups demonstrate that MAEO can operate correctly and stably under heterogeneous workloads as well as under controlled, uniform ones. Together, they confirm that MAEO’s implementation supports true hierarchical parallelism, enabling scalable deployment on modern HPC systems.

\section{Performance Benchmark and Applications}
\label{sec:opt_verif}

This section presents three sequential evaluations of the MAEO framework, each designed to assess a different aspect of its performance and applicability:

\begin{itemize}
\item \textbf{Single-problem benchmark:} Demonstrates the internal behavior of MAEO by analyzing the evolution of island HV values and population distributions during optimization.
\item \textbf{Wilcoxon comparison:} Performs \textit{pairwise} Wilcoxon signed rank tests between MAEO and each standalone MOO algorithm separately, quantifying relative convergence and diversity performance across benchmark problems.
\end{itemize}

Together, these evaluations provide a comprehensive view of MAEO’s mathematical performance, robustness, and potential for real-world applications.

\subsection{Single-problem Benchmark}
\label{sec:single_problem}

In this subsection, the MAEO framework is applied to a multiobjective benchmark problem to examine its optimization behavior. The variation of population size and HV across cycles, as well as both short-term and long-term migration dynamics are analyzed. The final Pareto front obtained from the optimization is also presented.

The chosen benchmark function for this test is the DTLZ2 problem, which features a smooth, continuous spherical Pareto front. It is commonly used as a baseline for evaluating diversity preservation in multiobjective algorithms. Due to its balanced complexity, clear front geometry, and suitability for visualization, DTLZ2 provides an effective platform for assessing MAEO’s convergence and diversity characteristics.

MAEO is configured with four optimization islands, each running a distinct multiobjective algorithm: NSGA-III, CTAEA, AGEMOEA2, and SPEA2. These algorithms are labeled as Island 1-4. \textit{Note that the algorithms assigned to each island in this study were chosen for demonstration purposes, and MAEO’s flexible design allows the integration of other MOO variants.}

\begin{figure}[H]
\centering
\includegraphics[width=0.8\linewidth]{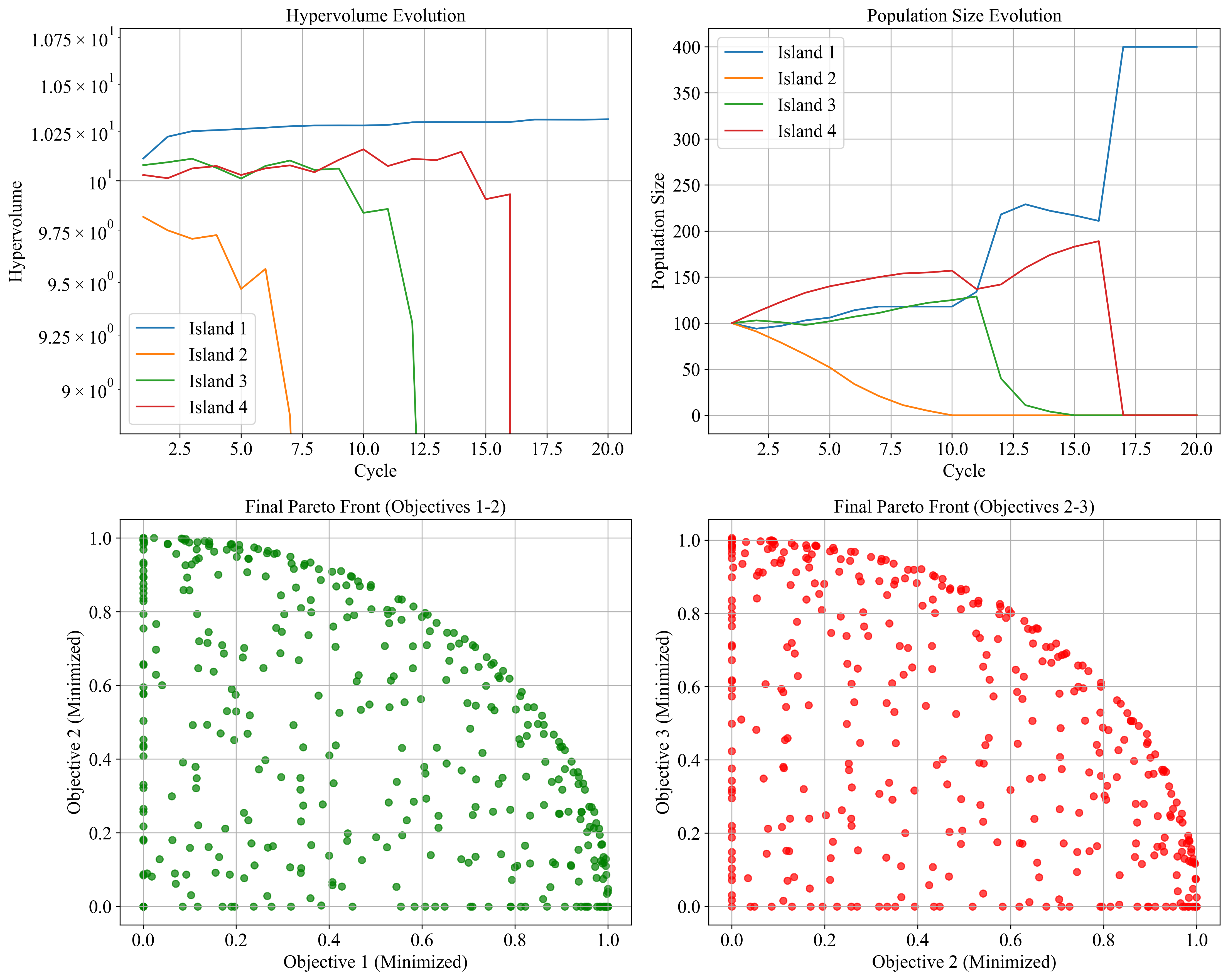}
\caption{MAEO performance on DTLZ2 benchmark function with 10 input parameters and 3 objectives. The reference point is set to [2.388, 2.238, 2.0350] by MAEO, resulting in an ideal Pareto-front hypervolume of 10.8757 in the case of an ideal solution. The islands 1-4 represent the algorithms NSGA-III, CTAEA, AGEMOEA2, and SPEA2, respectively.}
\label{fig:maeo_dtlz2}
\end{figure}

Each island is initialized with a population of 100 individuals. The NSGA-III island uses a crossover probability of 0.65 and a mutation probability of 0.35, applying a blending crossover scheme and logarithmic non-dominated sorting to preserve diversity within the evolving population. The remaining islands—SPEA2, AGEMOEA2, and CTAEA—use a simulated binary crossover with a distribution index of 20, which restricts offspring to remain close to their parent solutions and enhances convergence. A polynomial mutation operator with a distribution index of 20 introduces small, controlled perturbations to maintain exploration without destabilizing convergence. The crossover probability is set to 0.9, and the mutation probability is defined as the reciprocal of the number of decision variables, ensuring that, on average, one variable mutates per offspring. CTAEA additionally generates energy-based reference directions to support its two-archive selection process, while AGEMOEA2 and SPEA2 use their own adaptive archiving and environmental selection schemes to balance diversity and convergence. Each island operates independently and exchanges individuals through MAEO’s adaptive migration mechanism, promoting both broad exploration and efficient convergence across algorithms.
MAEO is run for 20 cycles with 10 generations in each cycle. Convergence assist is set to 0.2, meaning that first 16 cycles are used for migration, while the last 4 are used for convergence of the best-performing algorithm. The population distributions, island performances and the final population are given in Figure \ref{fig:maeo_dtlz2}.

The evolution of population size demonstrates that the MAEO framework performs as designed. Underperforming islands gradually lose individuals over successive cycles as their relative performance declines. These islands continue to receive a limited number of migrants from other islands in an attempt to recover and locate improved solutions. However, if an island fails to achieve a higher normalized performance score ($g_i$) than the next lowest island, its population is progressively depleted, and the island is deactivated—removing that optimization algorithm from further participation in MAEO. The remaining islands are then subjected to the same selection mechanism throughout the subsequent cycles.

The next observation concerns the “run-away” behavior in the population evolution. Once an island starts producing higher HV values, migration reinforces its advantage and it rapidly accumulates individuals. This run-away effect is deliberate: it allows MAEO to identify the best-performing algorithm and let it shape the final Pareto front. Normalizing HV by population size or using HV improvement instead of raw HV prevents this mechanism from working, suppressing the dominant island and weakening convergence. For this reason, unnormalized HV is retained as the most effective driver of island selection.

In this configuration, Islands 2, 3, and 4 are deactivated at cycles 10, 15, and 17, respectively. The effect of the convergence-assist mechanism is evident in this process. At the end of cycle 16, the HV of Island 4 falls below that of Island 1, triggering the transfer of its remaining individuals to Island 1. The final four cycles are then executed exclusively on Island 1, explaining the sharp reduction in Island 4’s population at cycle 17. As described in Section~\ref{sec:conv_assist}, executing the final cycles on a single island is an intentional feature of the convergence-assist mechanism, ensuring sufficient solution diversity and eliminating redundant points on the final Pareto front.

The best-performing island, NSGA-III on Island 1, began with a HV value comparable to those of Islands 3 and 4 but quickly surpassed them. As its HV increased, it attracted migrated individuals from the weaker islands and ultimately generated the final Pareto front.

The observed initial decrease in the number of individuals in Island 2 could be interpreted as a potential limitation in the algorithm’s formulation, suggesting that the islands might require a longer development phase before inter-algorithm migration begins. However, this decline is mitigated by the definition of the binomial sampling success probability for individual removal ($h_i$), which dampens excessive population loss. To further investigate this behavior, an additional test was conducted by extending the initialization period—allocating 20\% of the total cycle budget to the island evaluation phase before migration commenced. The resulting migration dynamics, however, remained nearly identical to those shown in Figure~\ref{fig:maeo_dtlz2}. This outcome indicates that island failure is not directly caused by the migration formulation but rather by two underlying factors: the stochastic nature of island initialization and the intrinsic performance of each optimization algorithm on the given problem.  

\begin{figure}
\centering
\includegraphics[width=0.6\linewidth]{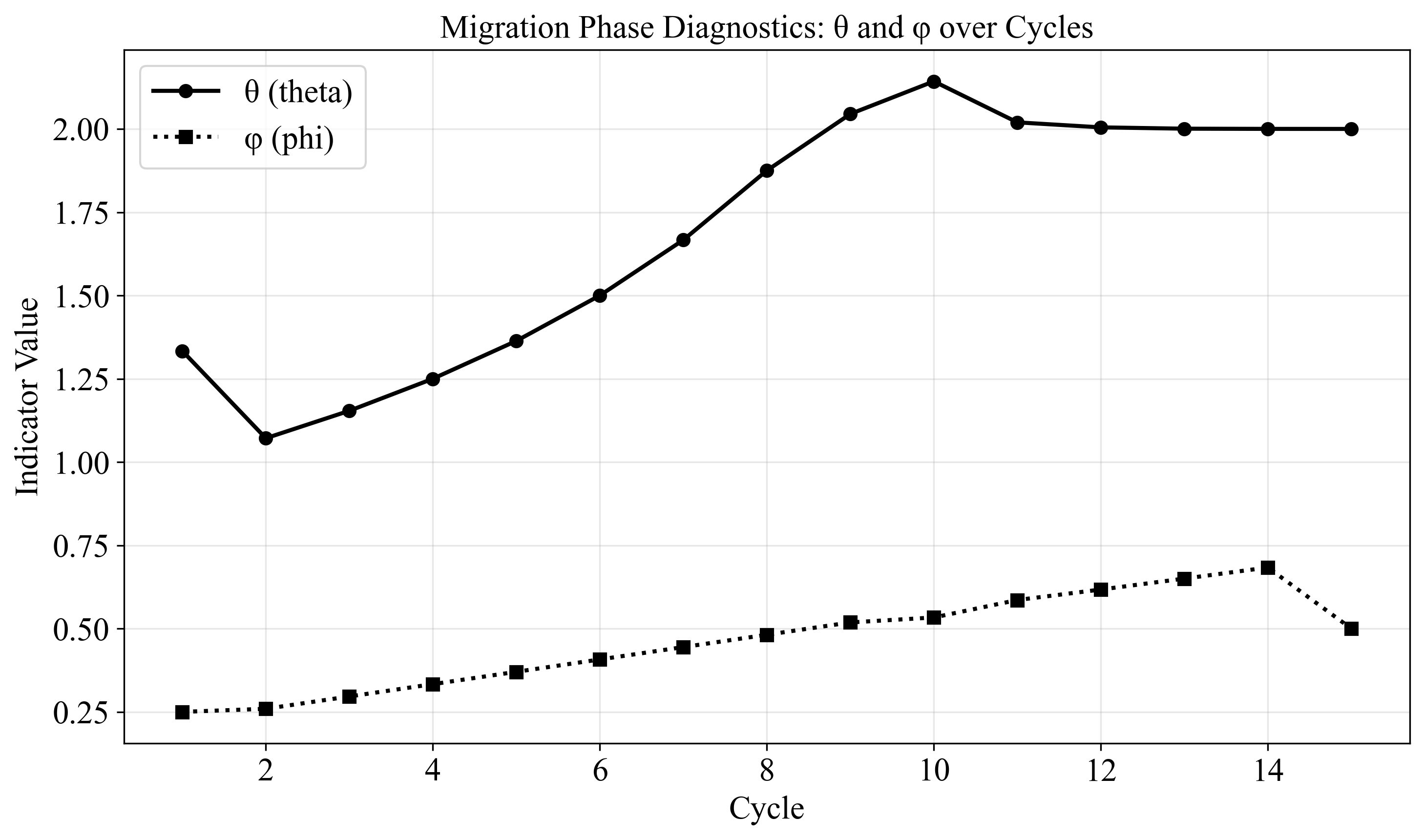}
\caption{MAEO performance on DTLZ2 benchmark function with 10 input parameters and 3 objectives.}
\label{fig:maeo_theta_phi}
\end{figure}

Migration characteristics follow the formulation provided in Section~\ref{sec:migration}. The short-term mobility indicator $\theta$ (heuristic speed indicator) and the long-term mobility indicator $\phi$ (average mobility index) are shown in Figure~\ref{fig:maeo_theta_phi}.

Migration indicators and the individual-removal probability are defined in Section~\ref{sec:migration} to characterize how individuals are redistributed among islands. As the binomial sampling success probability for individual removal ($h_i$) increases with the number of cycles, both indicators $\theta$ and $\phi$ are expected to rise over time, reflecting enhanced migration activity. However, as weaker islands become deactivated, the likelihood that a migrating individual returns to its original island increases. This feedback effect leads to a reduction in the short-term mobility indicator $\theta$. In contrast, $\phi$, which characterizes the long-term convergence behavior of the mobility matrix, is less sensitive to temporary fluctuations and island deactivations, being primarily influenced only after the second island is disabled. 

It can be observed that even when only two islands remain active, the migration indicators retain higher values compared to the initial optimization cycles. This observation is significant for two main reasons. First, although the number of individuals in each island remains approximately constant during the early cycles, the islands continue to exchange individuals, facilitating information sharing and supporting weaker islands through migration. Second, the elevated short-term mobility indicator $\theta$ indicates a substantially higher probability of individual transfer than in the early stages of the algorithm. This intensified migration accelerates the depletion of poorly performing islands, concentrating the population within the most successful islands and promoting rapid convergence toward the optimal solution of the fitness function.

\subsection{Wilcoxon Signed-rank Test}
The Wilcoxon signed-rank test is a nonparametric statistical method used to determine whether the performance differences between two optimization algorithms are significant across multiple independent runs. Because it does not assume normality in the distribution of results, it is well suited for stochastic multiobjective algorithms whose outcomes often exhibit skewness or heavy tails. By applying the test to scalar quality indicators such as HV indicator and IGD, it becomes possible to quantitatively assess whether one algorithm consistently outperforms another rather than relying solely on mean or median values.

A detailed justification for the choice of HV and IGD as performance indicators is provided in Section~\ref{sec:pop_eval}. HV is used as the primary metric for the Wilcoxon analysis, with IGD included to separately assess convergence behavior.

Two benchmark function suites are selected to evaluate the performance of MAEO against well-established MOO algorithms reported in the literature. The specific details and characteristics of these function sets are described in the following section.

\subsubsection{Selected Multiobjective Benchmark Functions}

The two selected benchmark suites cover a broad range of structural difficulties to comprehensively evaluate the robustness and generality of the proposed optimization algorithm. The \textbf{ZDT} problems provide two-objective test cases with distinct Pareto-front geometries and decision-space properties. \textit{ZDT1} presents a simple convex front that verifies baseline convergence and diversity maintenance, \textit{ZDT2} introduces concavity to assess selection pressure adaptation, and \textit{ZDT3} contains disconnected Pareto segments that examine the algorithm’s ability to preserve diversity across multiple niches. \textit{ZDT4} adds pronounced multimodality in the decision space to reveal premature convergence tendencies, and \textit{ZDT6} applies a biased mapping with a non-uniform Pareto-front density that challenges reference-vector and crowding-based diversity mechanisms.

The \textbf{DTLZ} problems extend these challenges to scalable many-objective settings, incorporating varying front geometries and nonlinear transformations. \textit{DTLZ1} defines a linear front used to test convergence under scaling, \textit{DTLZ2} offers a smooth spherical front that serves as a baseline for diversity assessment, and \textit{DTLZ3} retains the same geometry while introducing multimodality, probing the ability to escape local optima. \textit{DTLZ4} biases the search toward extreme points in the objective space, exposing weaknesses in diversity operators, while \textit{DTLZ5} represents a degenerate, low-dimensional Pareto manifold that examines dimensionality reduction capability. \textit{DTLZ6} introduces nonlinear scaling of the distance function, challenging algorithms with variable coupling and search bias, and \textit{DTLZ7} provides a discontinuous and locally optimal non-dominated sets to test global exploration. 

Together, these twelve benchmark functions collectively span convex, concave, multimodal, disconnected, biased, degenerate, nonlinear, and locally optimal non-dominated sets across both low- and high-dimensional objective spaces. 

\subsubsection{Wilcoxon Signed-Rank Test Results}

MAEO is compared against seven standalone, non-ensemble MOO algorithms: SPEA2, CTAEA, NSGA-II, NSGA-III, RVEA, AGEMOEA2, and MOEA/D. Four of these algorithms are also incorporated within MAEO as the constituent optimization islands.

The Wilcoxon signed-rank test is applied to determine, with 95\% confidence $(\alpha = 0.05)$, whether each algorithm performs statistically significantly better or worse than MAEO. To obtain statistically reliable results, each algorithm configuration is executed 20 times using different random seeds.

Each selected benchmark function has a predefined number of objective dimensions—typically 3, 5, or 10—which are indicated for each row in the Wilcoxon results tables. In addition, every benchmark function is evaluated under three input-dimensional settings (10, 30, and 50 variables), producing a total of 36 distinct problem configurations. Within the tables, the first, second, and third characters in each cell correspond to the results for 10-, 30-, and 50-dimensional problems, respectively.

To ensure a fair comparison, all algorithms are allocated an equal total number of fitness evaluations—400 evaluations per generation and 200 generations. In the MAEO configuration, this corresponds to 100 evaluations per island per generation, distributed across four islands representing distinct optimization algorithms. The ensemble operates for 25 cycles, each comprising 8 generations. Consistent with the earlier experiments, the convergence-assist parameter is fixed at 0.2. The Wilcoxon test results using the HV indicator as the performance metric are presented in Table~\ref{tab:wilcoxon_results_HV}. The Wilcoxon test results using the IGD as the performance metric are presented in Table~\ref{tab:wilcoxon_results_IGD}.

\begin{table}
\centering
\caption{Wilcoxon signed-rank comparison results with HV indicator metric. Symbols represent statistical significance: “+” indicates MAEO is better, “–” indicates worse, and “0” indicates no significant difference. Results from the test pertaining to the 10, 30 and 50-dimensional versions of each function are arranged from left to right. The total ``+'' or ``-'' counts are bolded if they meet or exceed the corresponding opposite counts for the 10-, 30-, and 50-dimension tests.}
\label{tab:wilcoxon_results_HV}
\resizebox{\textwidth}{!}{
\begin{tabular}{l c ccccccc}
\toprule
\textbf{Function} & \textbf{Objs.} & \textbf{SPEA2} & \textbf{CTAEA} & \textbf{NSGA-II} & \textbf{NSGA-III} & \textbf{RVEA} & \textbf{AGEMOEA2} & \textbf{MOEA/D} \\
\midrule
DTLZ1 & 5  & +++ & + - - & 0 - - & 0 - - & + - - & + - - & + - - \\
DTLZ2 & 3  & +++ & - - - & +++ & +++ & - - - & - - - & + - - \\
DTLZ3 & 5  & 0++ & 0 - - & - - - & - - - & 0 - - & 0 - - & 0 - - \\
DTLZ4 & 10 & +++ & - - - & +++ & - - - & - - - & - - - & - 0 - \\
DTLZ5 & 3  & 0+0 & - ++ & +++ & +++ & +++ & - - - & - - - \\
DTLZ6 & 5  & +++ & +++ & 0 - - & + - - & + - - & +++ & +++ \\
DTLZ7 & 10 & +++ & +++ & +++ & - - - & +++ & +++ & +++ \\
ZDT1  & 2  & - - - & - 0+ & +++ & +++ & +++ & - - + & - - - \\
ZDT2  & 2  & 00 - & - ++ & +++ & +++ & +++ & - - + & - - - \\
ZDT3  & 2  & - - - & + - - & ++ - & ++0 & +++ & - - - & - - - \\
ZDT4  & 2  & - - - & - - - & - - - & - - - & - - - & - - - & - - - \\
ZDT6  & 2  & + - - & + - - & + - - & + - - & +0 - & + - - & + - - \\
\midrule
\textbf{Total “+”} &     & \textbf{6/7/6} & 5/4/5 & \textbf{8/7/6} & 7/5/4 & 8/5/5 & 4/2/4 & 5/2/2 \\
\textbf{Total “–”} &     & 3/4/5 & \textbf{6/6/7} & 2/5/6 & 4/7/7 & 3/6/7 & \textbf{7/10/8} & \textbf{5/9/10} \\
\bottomrule
\end{tabular}
}
\end{table}

\begin{table}
\centering
\caption{Wilcoxon signed-rank comparison results with IGD metric. Symbols represent statistical significance: “+” indicates MAEO is better, “–” indicates worse, and “0” indicates no significant difference.}
\label{tab:wilcoxon_results_IGD}
\resizebox{\textwidth}{!}{
\begin{tabular}{l c ccccccc}
\toprule
\textbf{Function} & \textbf{Objs.} & \textbf{SPEA2} & \textbf{CTAEA} & \textbf{NSGA-II} & \textbf{NSGA-III} & \textbf{RVEA} & \textbf{AGEMOEA2} & \textbf{MOEA/D} \\
\midrule
DTLZ1 & 5  & 0 - - & +++ & +++ & +++ & +++ & +++ & +++ \\
DTLZ2 & 3  & 00+ & +++ & - - - & - - - & +++ & +++ & - -+ \\
DTLZ3 & 5  & 0+0 & +++ & +++ & +++ & +++ & +++ & +++ \\
DTLZ4 & 10  & - - - & +++ & -0+ & +++ & +++ & +++ & +++ \\
DTLZ5 & 3  & +++ & + - - & - - - & - - - & - - - & +++ & 0++ \\
DTLZ6 & 5  & - - - & - - - & - ++ & +++ & - - + & - - - & - - - \\
DTLZ7 & 10  & - - - & -0+ & 0++ & +++ & -++ & - - - & - - - \\
ZDT1  & 2  & +++ & +0 - & - - - & - - - & - - - & ++ - & +++ \\
ZDT2  & 2  & +++ & + - - & - - - & - - - & - - - & ++ - & +++ \\
ZDT3  & 2  & +++ & - ++ & - - - & - - - & - - - & +++ & - ++ \\
ZDT4  & 2  & +++ & +++ & +++ & +++ & +++ & 000 & +++ \\
ZDT6  & 2  & - ++ & - ++ & - ++ & - ++ & - ++ & - ++ & +++ \\
\midrule
\textbf{Total “+”} & & \textbf{6/6/8} & \textbf{8/7/8} & 3/6/7 & \textbf{6/7/7} & 5/6/8 & \textbf{9/9/8} & \textbf{7/9/10} \\
\textbf{Total “–”} & & 4/4/4 & 3/3/4 & 8/5/5 & 6/5/5 & 7/5/4 & 3/2/4 & 4/3/2 \\
\bottomrule
\end{tabular}
}
\end{table}

The Wilcoxon test results reveal meaningful differences between performance metrics when comparing MAEO to other MOO algorithms. Using the HV indicator (Table~\ref{tab:wilcoxon_results_HV}), MAEO performs better than two algorithms, is statistically equivalent to two, and worse than three. When evaluated with the IGD metric (Table~\ref{tab:wilcoxon_results_IGD}), MAEO performs better than five algorithms and equivalent to two. 

The number of objectives in each benchmark function does not significantly influence these outcomes. Mixed behavior is visible across functions with 3, 5, and 10 objectives. In contrast, the number of input dimensions has a noticeable effect on both HV- and IGD-based comparisons. As shown in the fifth and sixth columns of Table~\ref{tab:wilcoxon_results_HV}, MAEO generally achieves higher (better) HV than NSGA-II and NSGA-III when the decision space is limited to 10 variables; however, this advantage diminishes as dimensionality increases. Conversely, Table~\ref{tab:wilcoxon_results_IGD} shows that MAEO tends to have higher (worse) IGD than NSGA-II at 10 variables, indicating that \textbf{MAEO generates a more diverse but slightly less converged Pareto front in low dimensional decision spaces}; this effect also disappears as the dimensionality grows.

The grouping patterns of “–” symbols across rows reflect the strong problem dependence of MAEO’s performance. For example, MAEO consistently attains lower HV values on DTLZ2 and ZDT4, but performs strongly on DTLZ6 and DTLZ7. Importantly, when MAEO has lower HV on DTLZ2 and ZDT4, Table~\ref{tab:wilcoxon_results_IGD} shows that it simultaneously achieves significantly lower IGD, indicating superior convergence despite reduced HV. Conversely, in problems where MAEO achieves high HV (DTLZ6 and DTLZ7), the IGD comparisons become balanced, suggesting that its strong diversity performance is accompanied by acceptable convergence levels. Overall, the HV-based Wilcoxon tests show MAEO performing comparably to other algorithms, while the IGD-based tests highlight that MAEO tends to converge more effectively across a majority of benchmark problems.

These results also illustrate the underlying philosophy of MAEO: identifying the best-performing algorithm for a given problem and reallocating resources toward its island. Since MAEO’s ensemble uses NSGA-III, CTAEA, AGEMOEA2, and SPEA2, the ideal behavior is for the ensemble to detect the strongest island and progressively reinforce it. The IGD results for DTLZ1, DTLZ4, ZDT4, and ZDT6 exemplify this behavior—MAEO outperforms all methods except the top-performing algorithm in each respective test (SPEA2 in DTLZ1 and DTLZ4, AGEMOEA2 in ZDT4, and CTAEA in ZDT6). These cases demonstrate \textbf{MAEO’s ability to identify the most promising algorithm for the problem and to focus computational resources accordingly}, even achieving convergence results close to the best standalone method in reduced time, as seen in ZDT6 for CTAEA.

The island configuration used in MAEO directly influences these Wilcoxon test outcomes. The chosen set of four islands is designed to provide diversity while preserving sufficient computational budget for each island to develop. Increasing the number of islands would dilute the evaluation budget and diminish per-island learning capacity, while reducing the number could decrease the diversity needed for effectively handling different problem classes. Although this configuration may introduce minor biases favoring the four selected algorithms, the Wilcoxon results indicate that such effects are limited—particularly in comparisons involving RVEA and MOEA/D.

Finally, the convergence-assist mechanism plays an important role in determining MAEO’s standing in HV and IGD comparisons. If MAEO selects an island that prioritizes diversity, the resulting Pareto front may have excellent spread but lower HV, causing MAEO to be outperformed in HV-based evaluations. Conversely, if the island with the strongest convergence characteristics holds the highest HV at the end of the migration phase, MAEO can match or exceed the performance of the top standalone method for the given benchmark. This behavior reflects MAEO’s design goals: \textbf{balancing exploration and convergence during early stages, and concentrating computational effort on the empirically best-performing algorithm in later cycles}.

\section{Engineering Application: Equilibrium Cycle Optimization of a Small Modular Reactor (SMR) Core Design}
\label{sec:nuclear}

An important component of this study is the application of MAEO to a complex, real-world engineering problem. MAEO is designed for black-box optimization of computationally expensive models, making it well-suited for nuclear reactor design tasks. To assess its practical capability, MAEO is applied to the equilibrium-cycle optimization of an SMR design. The selected problem presents conflicting objectives and design constraints.

\subsection{Problem Definition}

\begin{figure}
\centering
\begin{minipage}{0.31\linewidth}
\centering
\includegraphics[width=\linewidth]{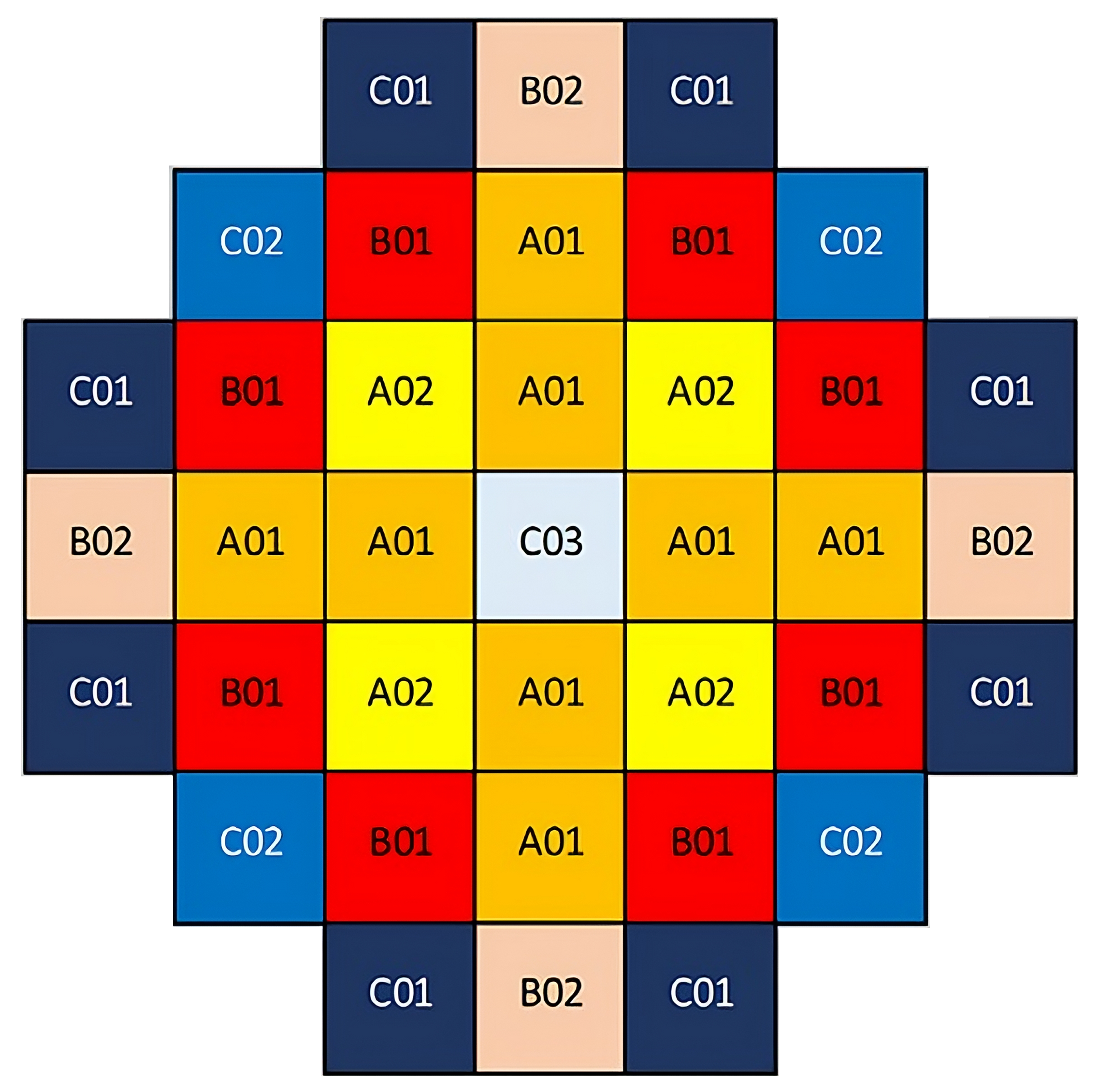}
\end{minipage}
\hfill
\begin{minipage}{0.4\linewidth}
\centering
\includegraphics[width=\linewidth]{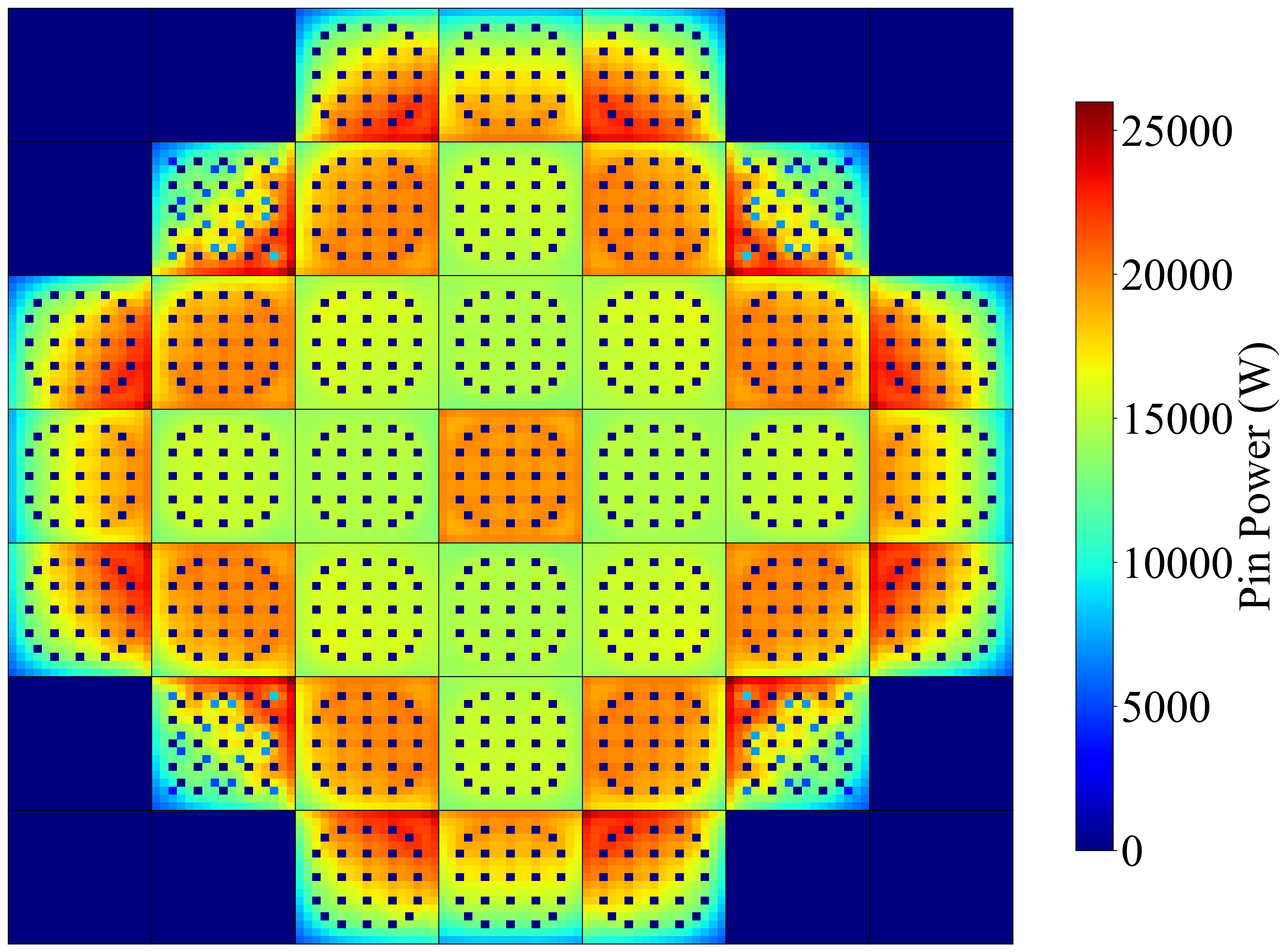}
\end{minipage}
\caption{Assembly types in the SMR-like reference core based on NuScale Power design, and the SIMULATE3 pin-power distribution of the reference core at BOL under cold zero power conditions.}
\label{fig:nuscale_pinpow}
\end{figure}

The SMR-like core configuration from Fridman et al.~\cite{fridman2023nuscale} is adopted as the reference design, which is based on the design made by the NuScale Power company. As illustrated in Figure~\ref{fig:nuscale_pinpow}, the core consists of seven fuel assembly types—A01, A02, B01, B02, C01, C02, and C03—with nominal U-235 fuel enrichments of 1.5\%, 1.6\%, 2.5\%, 2.6\%, 4.05\%, 4.55\%, and 2.6\% U-235 by weight, respectively. Among these, only C02 assemblies incorporate burnable absorber material, consisting of 16 rods with 8\% Gadolinium Oxide (Gd$_2$O$_3$) loading.

The fuel shuffling of the core is done in 3 cycles. B assemblies are moved to the location of A assemblies, C assemblies are moved to the location of B assemblies, C assemblies are replenished in every fuel cycle. For example, A01 assembly positions are replaced by previously B01 assemblies, B01 assembly positions are replaced by previously C01 assemblies, C01 assembly locations are loaded with fresh fuel. The only exception to this rule is the location of the C03 assembly, which is replenished in the beginning of every cycle.

A deterministic full-core model is constructed in \textsc{SIMULATE3}. The exact materials, pin dimensions, control rod definitions, assembly dimensions, reflector definitions have been used from the paper by Fridman et al. \cite{fridman2023nuscale}. The simulated behavior aligns closely with the reference: the beginning of life (BOL) cold-zero-power (CZP) reactivity at 1000 ppm boron is computed as 2301 pcm (compared to 2768 pcm reported in the reference), and radial power distributions agree with published results. The pin power plots printed by SIMULATE3 are identical to the reported pin powers in the reference study, which is given in Figure \ref{fig:nuscale_pinpow}. SIMULATE3 and its companion lattice physics code CASMO4 \cite{rempe1989simulate} (for nuclear data library generation) are validated nuclear simulation codes used by the industry for nuclear reactor analysis, including SMR designs. 

The difference in core-average exposure between cycles 5 and 6 is 0.335\%, indicating that equilibrium behavior is effectively reached by the fifth cycle. Accordingly, all optimization objectives and constraints in this study are evaluated at the fifth-cycle equilibrium state. The operating metrics extracted from the 5th cycle \textsc{SIMULATE3} results are provided in Table~\ref{tab:nuscale_params}. \textit{Notably, the reference design exceeds the F$\Delta$H safety limit of 1.50 ((Enthalpy-Rise Hot-Channel Factor)), motivating the need for optimization}.

\begin{table}[H]
\centering
\caption{SIMULATE3 reference (before optimization) results of the SMR-like NuScale design over the first five cycles.}
\label{tab:nuscale_params}
\begin{tabular}{ccc}
\toprule
Parameter & Value & Notes \\
\midrule
F$_q$ & 2.293 & Minimize ($<$ 2.49) \\
F$\Delta$H & 1.570 & Minimize ($<$ 1.5) \\
Max. Boron (ppm) & 1265.2 & Minimize if possible \\
Equilibrium EFPY & 4.9114 & Maximize if possible \\
Equilibrium Fuel LCOE (\$/MWh) & 14.6349 & Minimize if possible \\
\bottomrule
\end{tabular}
\end{table}

\subsection{Decision Variables}
\label{sec:decision_variables}

In this study, the enrichment values of these seven assemblies and the burnable absorber configuration constitute the discrete design variables. To maintain compatibility with the reference design and avoid exponential growth of the combinatorial space, the three-batch fuel-shuffling scheme is kept fixed. Thus, the optimization focuses on enrichment adjustments and burnable absorber placement within the established shuffling framework.


Fuel enrichments are allowed to vary within $\pm 0.4$~wt\% of their nominal values using discrete steps of 0.02~wt\%. This step size is sufficiently fine to approximate a continuous design space while remaining compatible with the nuclear code numerics, which supports roughly 600 unique assembly definitions per library. With 41 enrichment levels for each of the seven assemblies, the discrete subspace of enrichment values contains $41^7$ possible configurations.

Burnable absorber placement constitutes the last decision variable. In the reference design, only C02 assemblies contain sixteen 8\% Gd$_2$O$_3$ rods. To expand the design space while respecting pin-level viability, four BA configurations are allowed: (i) no BA rods, (ii) all 64 rods placed in C01 assemblies, (iii) same number of rods split across C01 and C02 assemblies, and (iv) the same number of rods placed in C02 assemblies. The spatial pattern of rods within an assembly is kept identical to that of the reference core. These eight discrete decision variables yield a total design space of: $41^7 \times 4 \;\approx\; 7.79\times 10^{11}$ configurations.

\subsection{Objectives and Constraints}
\label{sec:objectives}

Three objectives are selected to reflect key economic and physical performance indicators of an SMR core:  
(1) fuel levelized cost of electricity (LCOE),  
(2) maximum soluble boron concentration, and  
(3) cycle length expressed in fuel effective full power years (EFPY) of reactor operation.  
These objectives are intentionally conflicting and generate a meaningful Pareto surface for MOO.

\textbf{Fuel LCOE:} LCOE integrates both economic and reactor-physics variables and therefore provides a comprehensive measure of core-level fuel cost performance. Numerous formulations exist in the literature, differing in their treatment of capital recovery, discounting, outage structures, and fuel-cycle cost components (e.g., enrichment price, fabrication, disposal). Recent works demonstrate that fuel-cycle choices and irradiation performance can meaningfully influence LCOE estimates for both microreactors~\cite{Abdusammi2025Microreactors} and small modular reactors~\cite{asuega2023techno}. In this study, the fuel-cycle-focused LCOE formulation proposed by Seurin and Shirvan~\cite{SeurinShirvan2024FuelOptimization} is used. \textbf{Minimizing the LCOE} as equilibrium-cycle performance directly translates to lower per-assembly costs, and higher final assembly burnups. The explicit formulation and all supporting parameter definitions are provided in Section~\ref{sec:LCOE}. 

\textbf{Maximum Soluble Boron Concentration:} The second objective is the peak soluble boron concentration observed in the first 5 cycles. High boron concentration accelerates boron deposition in upper core regions, pushing the axial power shape toward excessive axial offset. Additionally, high boron levels degrade moderator temperature coefficient (MTC) and moderator void coefficient (MVC), potentially producing positive feedbacks that are prohibited by safety requirements.  Typical nuclear reactors constrain peak boron concentrations to roughly 1500--1600~ppm. In the reference SMR-like design, the maximum first-cycle concentration is 1099~ppm. \textbf{Minimizing the peak boron concentration} improves the feedback characteristics of the core while competing directly with the high-burnup, low-cost fuel strategies that raise boron requirements.

\textbf{Fuel EFPY:} The third objective is equilibrium-cycle EFPY, a key measure of fuel utilization. EFPY expresses the full-power-equivalent irradiation time of an assembly, standardizing its exposure independent of operational schedule. Because the SMR-like core uses a three-batch fuel scheme, the fuel EFPY is approximated as three times the equilibrium-cycle length. \textbf{Maximizing the EFPY} promotes high burnup and extended cycle lengths but conflicts with both boron reduction and the safety constraints.

Finally, F$_q$ (Heat Flux Hot-Channel Factor) and F$\Delta$ H (Enthalpy-Rise Hot-Channel Factor) impose critical thermal and safety limits on the reactor core. A hard-constraint formulation is adopted for these parameters. The designs that breach F$_q$ and F$\Delta$H limits of 2.49 and 1.5 are eliminated from the optimization using heavy penalties. 

\subsection{LCOE Calculation}
\label{sec:LCOE}

The LCOE has been studied in many nuclear engineering works, and definitions vary substantially in which technical and economic terms are included. Some analyses focus largely on plant‐level cost drivers: overnight capital cost, financing assumptions, fixed and variable operational and maintenance (O\&M), capacity factor, plant lifetime. Others extend into core‐level or fuel‐cycle‐level detail: fuel enrichment, burn‐up, refuelling intervals, tail‐blowdown enrichment, spent fuel handling, as well as constraints arising from thermal/hydraulic or safety limits. For example, Halimi et al. investigate how core design, fuel costs, and spent fuel influence economics in SMRs, emphasizing that fuel‐cycle and burnup can shift LCOE estimates significantly \cite{halimi2024scale}. Another recent study on nuclear microreactors by Abdusammi, Khaleb, Gao, \& Verma uses optimization under uncertainty over reactor capacity, enrichment, tail enrichment, refuelling interval, and discharge burnup in addition to capital and O\&M costs to identify configurations minimizing LCOE \cite{Abdusammi2025Microreactors}. These examples illustrate that LCOE is not a uniform measure — different studies embed different assumptions about both reactor physics and financial parameters.

In the selected study conducted by Seurin and Shirvan, the LCOE of nuclear power plant designs has been calculated by a constructed equation considering the core parameters. This equation uses: EFPY, levelization period (T$_{lev}$), thermal efficiency ($\eta$), effective capacity factor (K$_f$), discount rate ($r$), burnup of each assembly at the end of cycle (Bu$_i$), overnight fuel-cycle cost per assembly (C$_{ik}$), ratio of each assembly type in the core configuration ($\alpha_i$), and wet annular burnable absorber cost (C$_{i,\text{waba}}$). Additionally, calculating EFPY requires: T$_{lev}$, K$_f$. Calculating K$_f$ requires: availability factor (K$_{av}$), refueling outage duration (tfo), maintenance outage duration (tmo). Calculating T$_{lev}$ requires: cycle duration (T$_{cy}$), number of fuel batches (n$_{batches}$). Calculating C$_{i,waba}$ requires: total number of wet annular absorber (WABA) units ($W_{ab}$), core mass (M$_{core}$), fabrication period (T$_{fabric}$). Along with the WABA cost, a constant waste storage cost (C$_{waste}$) is added to the final LCOE value. The LCOE is calculated using the formulation provided by Seurin and Shirvan \cite{SeurinShirvan2024FuelOptimization}, reproduced in Eq.~\eqref{eq:LCOE}.

\begin{equation}
\begin{gathered}
LCOE = 
\frac{EFPY(T_{\text{lev}})}
{\eta \times K_f(T_{\text{lev}}) \times 24}
\times 
\frac{r}{1 - \exp^{-r T_{\text{lev}}}}
\sum_{i=1}^{N_{FA}} 
\alpha_i \frac{1}{B u_i}
\sum_{ik} C_{ik} \\[1em]
K_f(T_{\text{lev}}) = K_{av} \left( 1 - \frac{T_{FO} + T_{MO}}{T_{\text{lev}}} \right) 
\quad , \quad
T_{\text{lev}} = \frac{T_{cy} \times n_{batches}}{365.25} \\[1em]
C_{ik} = c_{ik} \, \exp^{-r T_{ik}} 
\quad , \quad
C_{i,\text{waba}} = W_{ab} \times c_{i,waba} \times \exp^{-r T_{fabric}} \frac{1}{M_{core}} 
\quad , \quad
\text{EFPY} = K_f \times T_{lev}
\end{gathered}
\label{eq:LCOE}
\end{equation}

The Bu$_i$ and T$_{cy}$ information are extracted from the SIMULATE3 results. M$_{core}$, n$_{batches}$, and $\alpha_i$ are constant values taken from the NuScale SMR core design in the reference study. The parameters K$_f$, T$_{lev}$, EFPY, C$_{ik}$, C$_{i,waba}$ are calculated from the existing parameters. The remaining parameters are approximated from the typical nuclear power plant parameters. The list of the approximated parameters and their values are given in Table \ref{tab:typical_params}.

\begin{table}[h!]
\caption{Typical SMR parameters used in the LCOE calculation.}
\centering
\begin{tabular}{cc}
\toprule
Parameter & Value \\
\midrule
$M_{\text{core}}$ (kg) & 9700 \\
$T_{\text{fabric}}$ (years) & 1 \\
$c_{i,\text{waba}}$ & 1500 \\
$\eta$ & 0.33 \\
$K_{av}$ & 0.9 \\
$T_{FO}$ & $\tfrac{30}{365}$ \\
$T_{MO}$ & $\tfrac{20}{365}$ \\
$r$ & 0.045 \\
$C_{\text{waste}}$ & 0.001 \\
$n_{\text{batches}}$ & 3 \\
$\alpha_i$ & $\left(\tfrac{8}{37}, \tfrac{4}{37}, \tfrac{8}{37}, \tfrac{4}{37}, \tfrac{8}{37}, \tfrac{4}{37}, \tfrac{1}{37}\right)$ \\
\bottomrule
\end{tabular}
\label{tab:typical_params}
\end{table}

The nuclear reactor fuel cost calculator \cite{UxC_FuelCostCalc} provided by UxC, LLC was used to estimate the up-to-date fuel cost parameters ($C_{ik}$). The feed and tail assays are kept constant at 0.711\% and 0.25\% for all assemblies. The default values used for $U_3O_8$, conversion, UF$_6$, and separative work unit (SWU) costs are:  

\[
U_3O_8 = 63.45 \,\$/\text{lb U}_3\text{O}_8, \quad 
\text{Conversion} = 80 \,\$/\text{kgU as UF}_6, \quad
\]
\[
\text{UF}_6 = 251.25 \,\$/\text{kgU as UF}_6, \quad
\text{SWU} = 185 \,\$/\text{SWU}.
\]

Since the contribution of the initial operating cycles to the overall fuel LCOE becomes negligible in long-term operation, the equilibrium-cycle fuel LCOE calculation is performed using only the equilibrium-cycle exposure values of the discharged assemblies (A01, A02, and C03) while applying the fuel-cycle cost parameters of the fresh assemblies (C01, C02, and C03). The WABA cost term is excluded from the equilibrium-cycle LCOE because burnable absorber rods are used only in the first-cycle loading and are not present in subsequent batches, consistent with steady-state operation.

\subsection{Optimization Results}
\label{sec:nuscale_results}

The equilibrium-cycle optimization of the SMR-like core is performed using the MAEO configuration described previously, \textbf{consisting of four optimization islands running NSGA-III, CTAEA, AGEMOEA2, and SPEA2, referred to as Islands~1–4}. Each island uses the same algorithmic parameters given in Section~\ref{sec:single_problem}. The MAEO framework is executed for 20 cycles with 10 generations per cycle, and the convergence-assist parameter is set to 0.2. Accordingly, the final four cycles are dedicated exclusively to convergence on the best-performing island.

\begin{figure}
\centering
\includegraphics[width=\linewidth]{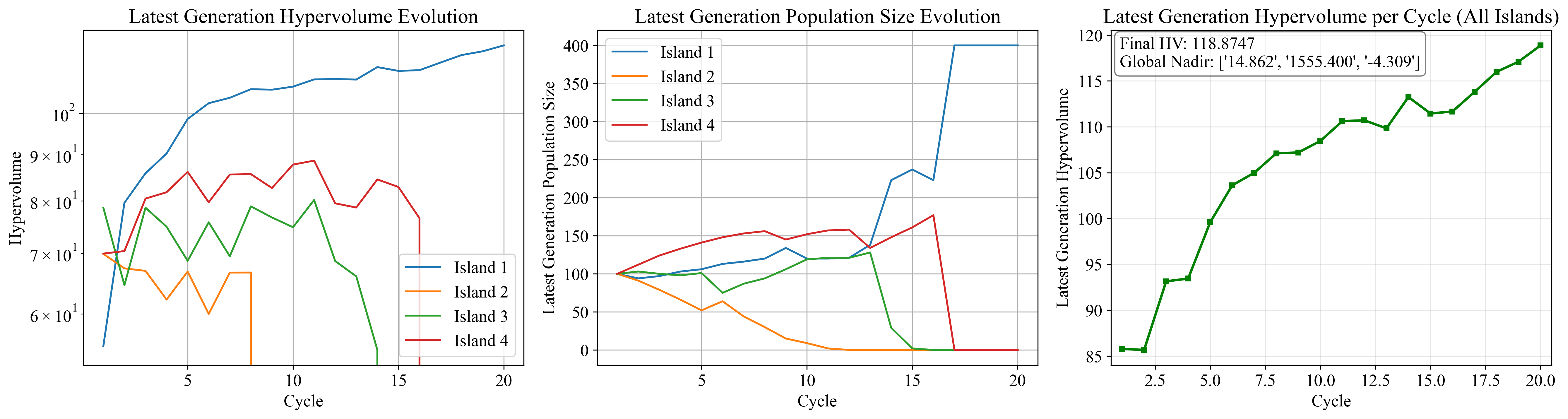}
\caption{Evolution of MAEO island performance during equilibrium-cycle optimization: Island HV values, population sizes, combined HV value across all islands. Each data point shows the final generation of each cycle.}
\label{fig:maeo_nuscale_stats}
\end{figure}

Figure~\ref{fig:maeo_nuscale_stats} presents the evolution of the four islands throughout the optimization. The left panel shows the HV value of each island at the last generation of each cycle. The middle panel tracks the population size of each island over cycles. The right panel displays the HV evolution of the combined population of all islands, summarizing the overall progress.

As shown in the HV and population-size evolution plots, attaining the highest HV value at a given cycle does not immediately translate into MAEO reallocating resources toward that island. For example, although NSGA-III reaches the highest HV by the end of cycle 2, its population does not become the largest until cycle 13. This behavior illustrates that MAEO deliberately allows all islands sufficient opportunity to develop before reallocating individuals based on their long-term performance. Once NSGA-III begins to exhibit a sustained increase in HV, migration progressively shifts more individuals toward it, and the weaker islands are disabled sequentially. The right-most panel, which reports the combined HV of all islands at each cycle, shows a smooth, monotonic increase. 

It is important to note that the reference point used in HV calculations is selected from the feasible, constraint-satisfying portion of the objective space. Designs that violate F$_q$ or F$\Delta$H constraints are excluded when determining the reference point, ensuring that HV comparisons reflect meaningful performance differences among feasible solutions.

Starting from cycle~16, no constraint violations are observed in any island. This behavior demonstrates that the convergence-assist mechanism performs as intended by concentrating all computational resources on the best-performing algorithm once the migration stage concludes. By shifting the algorithm from exploration to focused convergence, the remaining cycles avoid breaching safety limits, thereby eliminating unnecessary simulations.

\begin{figure}
\centering
\includegraphics[width=\linewidth,trim={0 0 0 0.8cm},clip]{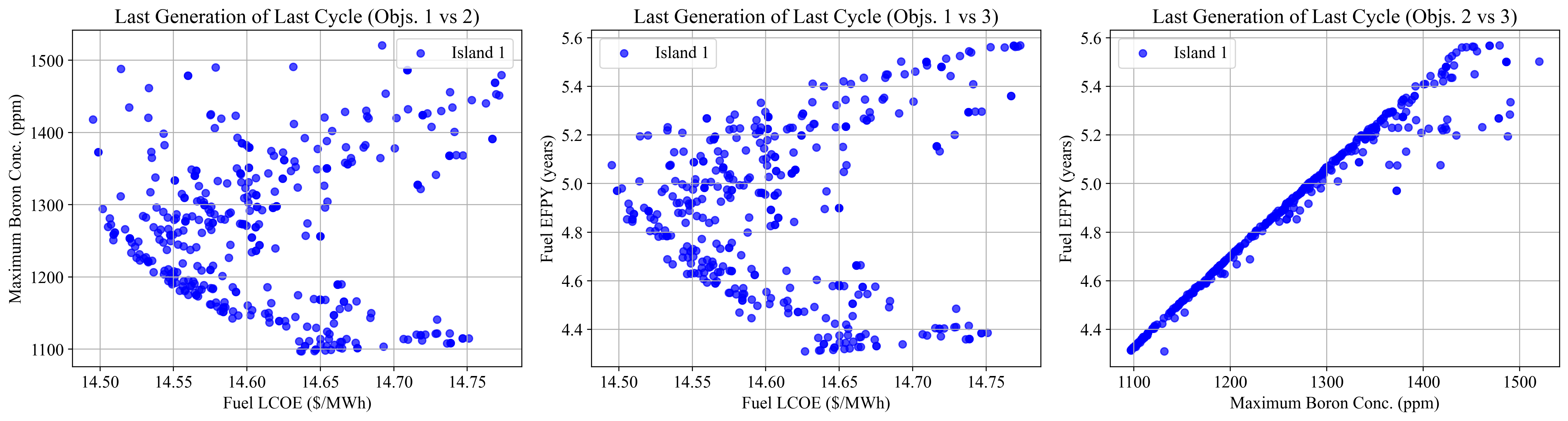}
\caption{Final Pareto-optimal solutions obtained at the last generation of the last MAEO cycle. Three pairwise objective projections are shown. All points satisfy F$_q$ and F$\Delta$H constraints.}
\label{fig:maeo_nuscale_last_gen}
\end{figure}

Figure~\ref{fig:maeo_nuscale_last_gen} shows the final Pareto-optimal designs obtained at the last generation of the last cycle. Pairwise objective projections are shown side-by-side. These plots represent the non-dominated solutions identified after MAEO completes its migration and convergence stages.

An important observation from these figures is the strong positive correlation between maximum boron concentration and fuel EFPY. Improvements in fuel EFPY (better values) coincide with increases in peak boron concentration (worse values), while reductions in boron concentration come at the expense of lower EFPY. In contrast, the relationship between fuel LCOE and other objectives allows a broader distribution of solutions spanning both high and low values, forming a trade-off surface. One might be tempted to select the design with the lowest LCOE and lowest boron concentration—visible in the right panel—as the “best” candidate. However, all solutions lie on a three-dimensional trade-off surface, and such a choice would necessarily yield a sub-optimal fuel EFPY. Identifying the most appropriate design therefore requires assessing the relative importance of all three objectives and selecting the solution that best aligns with the specific operational or economic priorities of the intended application.

\begin{figure}
\centering
\includegraphics[width=\linewidth]{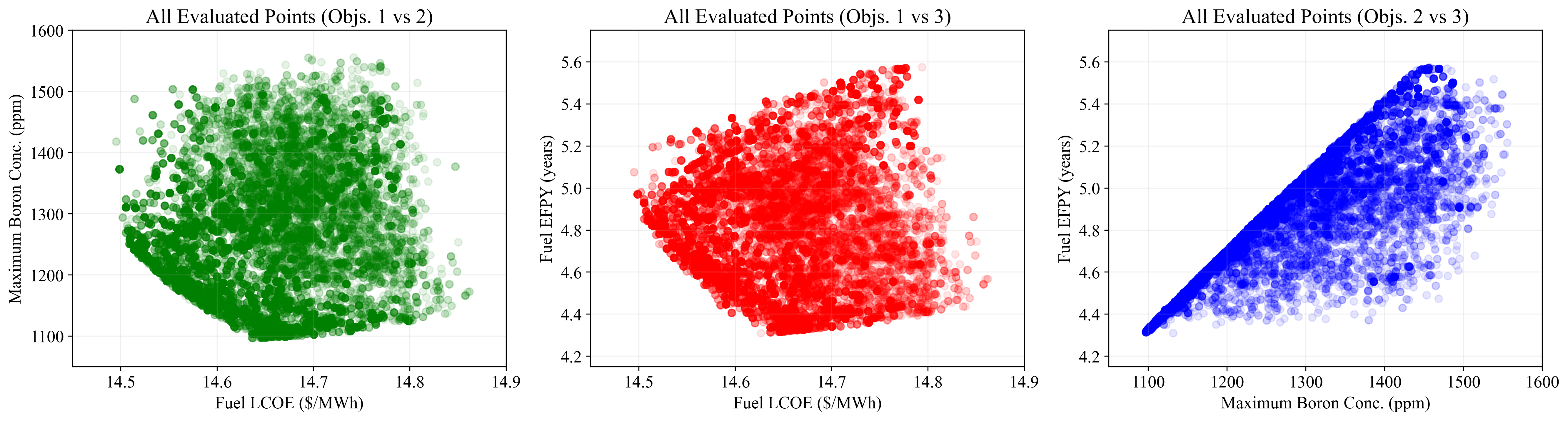}
\caption{All evaluated core designs from the MAEO optimization of the SMR-like core. The scatter plots show the sampled solutions in the same three objective projections, illustrating the explored design space and the effect of constraint filtering.}
\label{fig:maeo_nuscale_all_pts}
\end{figure}

Figure~\ref{fig:maeo_nuscale_all_pts} shows the distribution of all evaluated solutions sampled throughout the optimization process. These figures help illustrate the coverage of the design space and display the effect of the constraint handling. The algorithm’s exploration and the objective limits imposed by the constraints are clearly visible in this figure.

The physical design limits imposed by the constraints are clearly visible in all 3 panels. These limits define the boundaries of the Pareto front, as the optimizer cannot cross the regulatory thresholds set for the reactor core. 

In the left panel, the lower values of fuel LCOE and maximum boron concentration are clearly bounded by the F$_q$ and F$\Delta$H limits. A similar pattern is visible in the middle panel, where low-LCOE/high-EFPY combinations are truncated by the constraint surface. This behavior indicates that the feasible region enforces a strict limit on the best achievable objective values, preventing the optimizer from approaching otherwise ideal regions of the design space. The right panel further shows that designs with low maximum boron concentration and high fuel EFPY lie directly on the constrained boundary. Collectively, these results illustrate that all objectives are limited by the constraints, and the optimal solutions are expected to lie on the constraint surface.

Selecting a single core design from the explored equilibrium-cycle SIMULATE3 evaluations requires choosing a point in a three-dimensional objective space. Identifying a preferred design involves trade-offs among the three objectives, and the decision depends on how the decision-maker prioritizes them. In general, any weighting strategy may be adopted. 

For the purposes of this study, to provide a consistent and application-neutral reference and illustrate representative practical outcomes for the nuclear industry, all three objectives are assigned equal weight. Each objective dimension is first normalized to the 0–1 range, after which the scalarized objective value is computed as the sum of the three normalized components. The design with the smallest normalized objective sum is selected as the best-performing configuration in the explored population. The corresponding decision variables and objective values of this representative optimum are reported in Table~\ref{tab:best_cores}. To provide a diverse set of representative outcomes, a second design is identified by repeating the selection process with the two safety constraints. All five quantities (three objectives and two constraints) are independently normalized to the 0–1 range before computing the scalarized score. The best-performing design under this combined objective–constraint criterion is also reported in Table~\ref{tab:best_cores}.

\begin{table}[h!]
\centering
\fontsize{10}{12}\selectfont
\caption{Design variables and objective results for the best-performing MAEO-optimized SMR core configurations. The first row shows the solution with the minimum normalized objective-sum value, and the second row shows the solution with the minimum normalized objective-and-constraint-sum value. Note: BA refers to the assembly that constrains the burnable absorber rods.}
\label{tab:best_cores}
\begin{tabular}{lcccccc|c|ccccc}
\toprule
\multicolumn{7}{c|}{Enrichments (\%)} & \multicolumn{1}{c|}{} & \multicolumn{5}{c}{Objectives} \\
A01 & A02 & B01 & B02 & C01 & C02 & C03 & BA Loc. & LCOE & Boron & EFPY & F$_q$ & F$\Delta$H \\ 
\midrule
1.88 & 2.0 & 2.28 & 2.2 & 4.25 & 4.15 & 2.24 & C02 & 14.5020 & 1294.0 & 4.9795 & 2.166 & 1.490 \\
\midrule
1.88 & 2.0 & 2.54 & 2.38 & 4.05 & 4.15 & 2.52 & C02 & 14.5319 & 1375.8 & 4.8381 & 2.012 & 1.388 \\
\bottomrule
\end{tabular}
\end{table}

The rows of Table~\ref{tab:best_cores} demonstrate that it is possible to obtain core designs that improve fuel LCOE and maximum boron concentration at the expense of fuel EFPY, relative to the reference performance metrics listed in Table~\ref{tab:nuscale_params}. In the first design, the enrichment values of the low-enriched A and B assemblies increase, while those of the highly enriched C assemblies decrease. This redistribution reduces local power peaking and successfully brings the F$\Delta$H value below the regulatory limit.

When the safety constraints are incorporated directly into the normalized summation used for selecting the optimum, the optimizer shifts the core design more aggressively toward satisfying the constraint boundaries. In this second solution, enrichments in the low-enriched assemblies increase further, while enrichments in the highest-enrichment assemblies decrease even more. These adjustments degrade all three objective values compared to the first optimized design; however, they produce a more favorable safety profile, yielding lower F$_q$ and F$\Delta$H values.

In both optimized solutions, the burnable absorber (BA) configuration remains located in the C02 assemblies. This outcome is consistent with the reference NuScale core, where the highest F$_q$ and F$\Delta$H values occur in the C02 locations. Thus, retaining burnable absorber rods in these assemblies is a logical outcome of the optimization, as it prevents violation of the safety constraints. Although the optimizer explored alternative burnable absorber configurations, the final solutions shown in Figure~\ref{fig:maeo_nuscale_all_pts} correspond to the configuration in which all burnable absorber rods are placed exclusively in the C02 assemblies.

\section{Concluding Remarks}
\label{sec:conclusion}

This study introduced the Multiobjective Animorphic Ensemble Optimization (MAEO) framework, equipped with a two-level parallelism capability, developed to address the challenges of black-box, computationally expensive, and multiobjective design problems. MAEO combines multiple state-of-the-art evolutionary algorithms into a coordinated ensemble guided by hypervolume-based island performance metrics, adaptive migration, and a convergence-assist mechanism. The formulation aims to balance global exploration among heterogeneous solvers with focused convergence on the most promising algorithm as optimization progresses.

Benchmark studies on standard DTLZ and ZDT test suites demonstrated that MAEO provides competitive performance relative to standalone multiobjective algorithms. The Wilcoxon signed-rank analysis showed that MAEO delivers balanced hypervolume performance across diverse problem classes while achieving consistently strong convergence in terms of the inverse generational distance. The diagnostic analysis of migration behavior confirmed that the animorphic formulation successfully identifies the most suitable optimization algorithm for each problem, reallocates resources accordingly, and adaptively suppresses stagnating islands.

The real-world engineering application—equilibrium-cycle optimization of a nuclear small modular reactor core—illustrated the practical value of MAEO for large-scale engineering problems that integrate discrete design variables, expensive computer simulations, and conflicting physics-based objectives. The optimized designs showed measurable improvements in fuel-cycle economics, soluble boron concentration, and safety margins relative to the reference configuration. The study further demonstrated that MAEO can operate effectively within tight computational budgets, and that its convergence-assist strategy reduces unnecessary evaluations after the dominant island has emerged.

Overall, the results indicate that MAEO is a robust and flexible optimization approach capable of handling high-dimensional, multiobjective, mixed discrete–continuous, and simulation-intensive design problems. Its ensemble structure provides resilience against algorithmic weaknesses, while its animorphic migration logic allows it to adapt dynamically to the characteristics of each problem. Future work may focus on incorporating surrogate-assisted evaluations to the islands to make the algorithm oriented toward extremely expensive cost functions and extending the ensemble to include physics-informed or problem-specific optimizers.

\section*{Data Availability}
\label{sec:avail}

Currently, the authors possess, in a private GitHub repository, all the data and codes needed to reproduce the results in this work. To ensure confidentiality of this research, the authors will make this repository public during an advanced stage of the review process, and it will be listed under our research group's public GitHub page: \url{https://github.com/aims-umich}.

\section*{Acknowledgment}

This work is sponsored by the Department of Energy Office of Nuclear Energy's Distinguished Early Career Program (Award number: DE-NE0009424), which is administered by the Nuclear Energy University Program. 

\section*{CRediT Author Statement}

\begin{itemize}
    \item \textbf{Omer Faruk Erdem}: Conceptualization, Methodology, Software, Validation, Data Curation, Formal analysis, Visualization, Investigation, Writing - Original Draft. 
    \item \textbf{Dean R. Price}: Conceptualization, Methodology, Software, Writing - Review and Edit    
    \item \textbf{Paul Seurin}: Methodology, Software, Writing - Review and Edit    
    \item \textbf{Majdi I. Radaideh}: Conceptualization, Methodology, Funding acquisition, Supervision, Resources, Project administration, Writing - Review and Edit.
\end{itemize}

\bibliographystyle{elsarticle-num}
\setlength{\bibsep}{0pt plus 0.3ex}
{
\small
\bibliography{references}}

\end{document}